\documentclass{article}

\usepackage{microtype}
\usepackage{graphicx}
\usepackage{subcaption}
\usepackage{float}
\usepackage{booktabs}           
\usepackage{multirow}
\usepackage[most]{tcolorbox}
\usepackage{tcolorbox}
\usepackage{siunitx}
\usepackage{hyperref}
\usepackage{xcolor}

\usepackage[preprint]{icml2026}
\makeatletter
\renewcommand{\Notice@String}{}
\makeatother

\usepackage{amsmath}
\usepackage{amssymb}
\usepackage{mathtools}
\usepackage{amsthm}

\usepackage[capitalize,noabbrev]{cleveref}

\theoremstyle{plain}

\theoremstyle{definition}

\theoremstyle{remark}

\usepackage[textsize=tiny]{todonotes}

\icmltitlerunning{EO-Agents: A Three-Agent LLM Pipeline for Earth Observation Hypothesis Generation}

\begin{document}

\twocolumn[
\icmltitle{EO-Agents: A Three-Agent LLM Pipeline for Earth Observation Hypothesis Generation}

\begin{icmlauthorlist}
    \icmlauthor{Mahyar Ghazanfari}{gwu}
    \icmlauthor{Amin Tabrizian}{gwu}
    \icmlauthor{Armin Mehrabian}{nasa,adnet}
    \icmlauthor{Peng Wei}{gwu}
\end{icmlauthorlist}

\icmlaffiliation{gwu}{George Washington University, Washington, DC, USA}
\icmlaffiliation{nasa}{NASA Goddard Space Flight Center, Greenbelt, MD, USA}
\icmlaffiliation{adnet}{ADNET Systems, Inc., Bethesda, MD, USA}

\icmlcorrespondingauthor{Mahyar Ghazanfari}{mahyar.ghazanfari@gwu.edu}

\icmlkeywords{knowledge graphs, graph neural networks,
              hypothesis generation, LLM-as-judge,
              Earth observation, scientific discovery,
              multi-agent evaluation}

\vskip 0.3in
]

\printAffiliationsAndNotice{Accepted at ICML 2026 AI for Science Workshop.}

\begin{abstract}
Large language models have recently been explored for scientific hypothesis generation, but most prior work relies on unstructured literature and free-form textual claims. We present a pipeline for Earth observation that grounds hypothesis generation directly in the NASA Earth Observation Knowledge Graph. A heterogeneous graph neural network trained on historical co-usage relations ranks candidate dataset pairings, and a three-agent LLM pipeline filters, generates, and evaluates structured research hypotheses. Applied to 1,475 NASA datasets, the system produces 160 hypotheses spanning multiple Earth-science domains, including ecohydrology, glaciology, aerosol--cloud interactions, vegetation phenology, and stratospheric chemistry. Model-predicted novel dataset pairings are rated nearly as plausible as held-out real co-usages from the literature, indicating that the pipeline surfaces scientifically coherent yet unexplored combinations. A $2 \times 2 \times 2$ factorial experiment across GPT-5.2 and Claude Sonnet 4.6 shows that hypothesis rankings remain stable, while absolute scores depend strongly on judge identity, highlighting limitations of single-judge LLM evaluation. Code, dataset, and generated hypotheses can be found
\href{https://github.com/Mahyar-GH79/EO_Agent}{\textcolor{blue}{here}}.
\end{abstract}
\section{Introduction}

\begin{figure*}[!t]
    \centering
    \includegraphics[width=\linewidth]{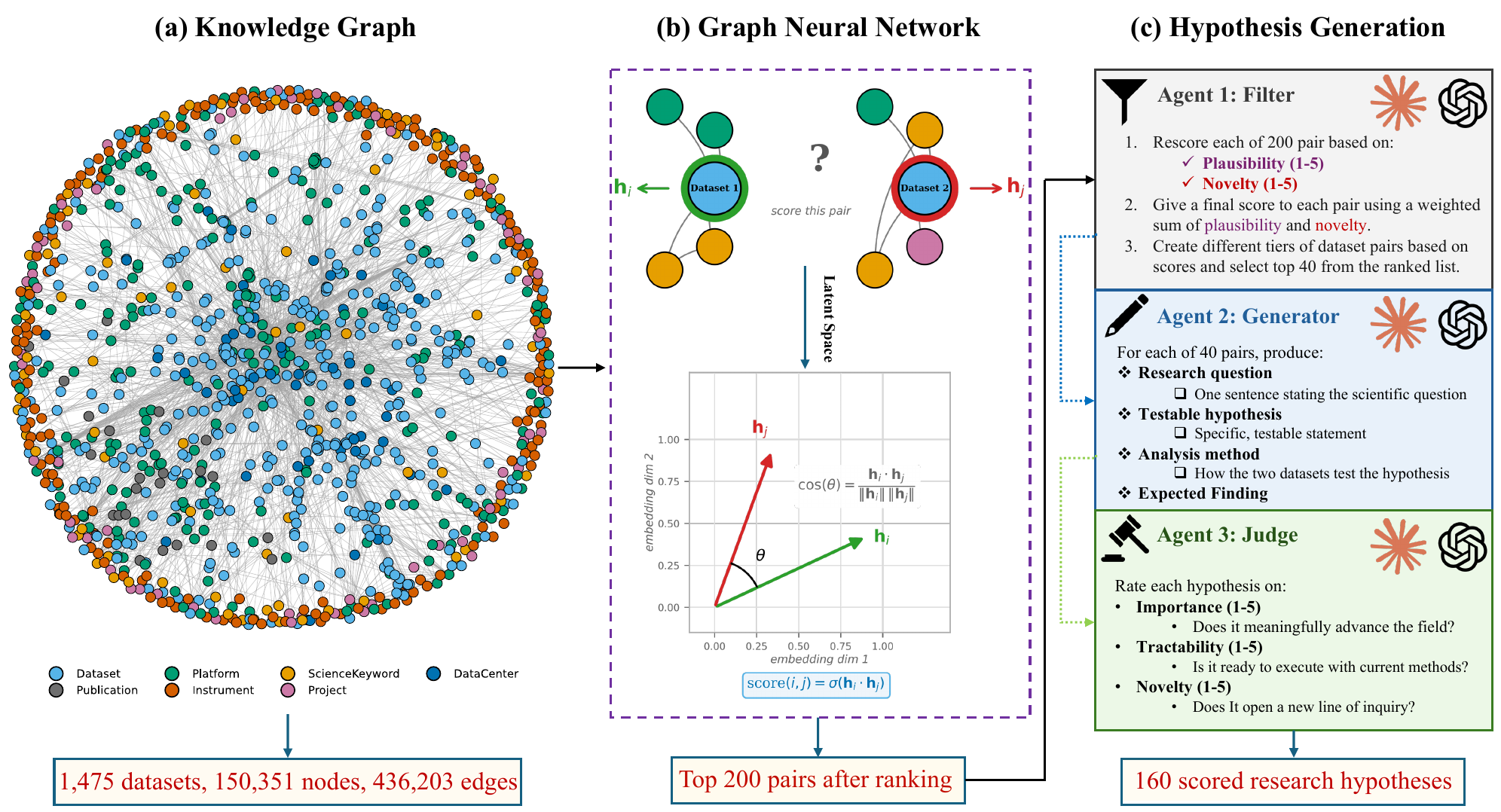}
    \caption{Three-panel overview of the pipeline.
    (a)~The NASA Earth Observation Knowledge Graph (150{,}351 nodes,
    436{,}203 edges across seven node types).
    (b)~A dense-hop neighborhood around a candidate dataset pair is
    passed through a heterogeneous GNN; node embeddings
    $\mathbf{h}_i,\mathbf{h}_j$ are compared with a dot-product scorer
    $\operatorname{score}(i,j) = \sigma(\mathbf{h}_i\cdot\mathbf{h}_j)$
    to produce a ranked list of 200 candidate pairs.
    (c)~Three LLM agents (each independently GPT-5.2 or Claude Sonnet~4.6)
    filter, generate, and judge hypotheses, yielding 160 scored
    research hypotheses.}
    \label{fig:pipeline}
\end{figure*}

Earth-observation (EO) research is fundamentally combinatorial. A
typical study fuses a soil-moisture retrieval with a vegetation
index, pairs a lidar canopy-height product with a global
digital-elevation model, or cross-references atmospheric chemistry
against stratospheric profiles. NASA alone distributes over
8{,}000 EO datasets across twelve archives, spanning hundreds of instruments and five decades of observations. The scientific payoff
is increasingly driven not by any single measurement but by the
\emph{pairing} a researcher chooses to study, and by whether that
pairing has been attempted before. The combinatorial space of
unordered pairings, ${\sim}10^{6}$ even under conservative
filtering, far exceeds what any individual researcher can survey,
and most of it remains scientifically unexplored.

Recent work has proposed large language models (LLMs) as engines
for scientific hypothesis generation, producing research ideas in
chemistry \citep{yang2025moose, yang2025moose2}, biomedicine
\citep{wang2024scimon}, materials science
\citep{ghafarollahi2024sciagents}, astrobiology
\citep{saeedi2025astroagents}, and machine-learning research itself
\citep{si2024canllms, lu2024aiscientist, yamada2025aiscientistv2}.
The dominant paradigm grounds LLM generation in unstructured
scientific text: a retriever surfaces related papers, a generator
composes an idea from their content, and an evaluator scores the
result \citep{baek2024researchagent, yu2025researchtown}. This
paradigm is effective for literature-driven domains where
hypotheses are themselves textual claims; in observational,
data-rich domains such as EO, however, a useful hypothesis is
inseparable from the specific measurement products that would test
it. Saying ``use satellite data to study drought'' is not a
hypothesis; saying ``combine SPL4SMGP soil moisture with MYD13Q1
vegetation indices to test whether 16-day EVI response depends on
biome-specific soil-moisture thresholds'' is. A literature-grounded
pipeline can in principle reach the second statement by extracting
dataset mentions from retrieved papers, but it must do so as a
side-effect of free-form text generation. We argue that grounding
the retrieval step directly in a typed knowledge graph of
measurement products is a more direct route, because every
generated hypothesis is then pinned to two named NASA datasets by
construction.

We therefore explore a complementary architecture in which the
retrieval step itself is structured. We retrieve over a
heterogeneous knowledge graph of EO research artifacts, where
datasets, platforms, instruments, science keywords, and projects
appear as typed nodes, and dataset co-usage (derived from joint
citation by a shared publication), platform--instrument mounting,
and dataset--metadata membership appear as typed edges. A
heterogeneous graph neural network, trained by link prediction on
past co-usages (publications through 2022), ranks candidate
dataset pairings by predicted co-usage likelihood and surfaces the
top 200 pairs not observed in the training, validation, or test
co-usage sets. A three-agent LLM pipeline then refines this list:
a \emph{filter} agent rescores the top candidates on plausibility
and novelty, a \emph{generator} agent articulates a structured
research hypothesis (question, testable claim, analysis method,
expected finding) for each surviving pair, and a \emph{judge} agent
rates the resulting hypothesis on importance, tractability, and
novelty under both blind and contextual conditions. \cref{fig:pipeline} shows the complete pipeline.

Applied to the NASA Earth Observation Knowledge Graph, the pipeline produces 160 hypotheses spanning multiple Earth science domains. Held-out real co-usages and model-predicted novel pairings receive comparable plausibility scores from LLM judges, suggesting that the system surfaces scientifically coherent yet unexplored combinations. A representative
example pairs IceBridge airborne magnetometer with InSAR-derived
Antarctic ice velocity: the generator proposes that subglacial
magnetic anomalies predict basal-friction patterns underlying
modern ice flow---linking solid-Earth geology to cryospheric
dynamics with public archives.

Because these scores can shift depending on which model plays which
role, we test the pipeline under all eight combinations of GPT-5.2
and Claude Sonnet 4.6 across the three agent roles. These eight
setups produce 640 judgments across the 160 hypotheses. We find
that the \emph{ranking} of hypotheses stays consistent (159 of 160
hypotheses change by at most 2 points across the four judge
conditions), but the \emph{absolute} scores depend heavily on which
model is the judge: about $25\%$ of the variation in importance
scores comes from the judge alone, versus less than $2\%$ from the
filter or the generator. For tractability, the biggest factor is
instead whether the judge sees the underlying datasets or only the
hypothesis text ($16\%$ of the variance). Together, these findings
suggest that relying on a single LLM judge is risky for absolute
scoring, and motivate the multi-judge protocol we use throughout
the paper.

Our contributions are threefold:
\begin{enumerate}
    \item We introduce a hypothesis-generation pipeline for
    observational science that ties each LLM-generated hypothesis
    to a specific pair of NASA datasets, surfaced by a
    knowledge-graph ranker rather than by free-form text retrieval,
    yielding 160 structured hypotheses across many Earth-science
    domains, with five flagship themes highlighted in
    \cref{app:flagship}.
    \item We show that the pairs the model predicts as novel are
    rated nearly as plausible by LLM judges as held-out real
    co-usages from the published literature, which is evidence that the
    pipeline produces scientifically coherent ideas beyond what the
    model has already seen.
    \item We run a $2{\times}2{\times}2$ factorial over filter,
    generator, and judge identity and find that, although
    hypothesis rankings are stable across setups, the absolute
    scores depend strongly on which model is the judge---a warning
    about relying on a single LLM judge for this kind of
    evaluation. We release the knowledge-graph preprocessing, the
    trained GNN, all 160 hypotheses, the 640 validator judgments,
    and the full analysis code.
\end{enumerate}

\section{Related Work}
\label{sec:related}

\paragraph{LLM hypothesis generation from literature.}
A growing line of work uses LLMs to generate research ideas grounded in
\emph{text}. Prior systems explore retrieval-based and multi-agent
approaches for scientific ideation, including SciMON
\citep{wang2024scimon}, ResearchAgent
\citep{baek2024researchagent}, MOOSE-Chem
\citep{yang2025moose}, and ResearchTown
\citep{yu2025researchtown}. Fully autonomous pipelines
\citep{lu2024aiscientist, yamada2025aiscientistv2} further extend this
to end-to-end ideation, experimentation, and manuscript writing, while
large-scale expert studies \citep{si2024canllms} show that
LLM-generated ideas can be judged more novel than human expert ideas,
though often less feasible. All of this work retrieves over
\emph{unstructured} scientific text and treats a hypothesis as a
free-form natural-language claim. In observational, data-rich domains
such as Earth observation, the unit of discovery is instead a specific
pair of measurement products whose joint analysis is executable, which
motivates our GNN-over-typed-KG ranker (\cref{sec:kg-gnn}) in place of
literature retrieval.

\paragraph{Multi-agent and data-grounded discovery.}
Closer to our setting, multi-agent pipelines ground generation in
structured artifacts. SciAgents \citep{ghafarollahi2024sciagents}
reasons over an ontological knowledge graph with Ontologist, Scientist,
and Critic roles for bioinspired materials; AstroAgents
\citep{saeedi2025astroagents} generates astrobiology hypotheses from
mass-spectrometry measurements with eight role-specialized agents;
MOOSE-Chem2 \citep{yang2025moose2} targets \emph{fine-grained}
hypotheses with concrete methodology via hierarchical search;
AutoDiscovery \citep{agarwal2025autodiscovery} drives open-ended
discovery across 21 real datasets with Bayesian surprise as an
LLM-defined reward under Monte Carlo tree search. Our pipeline shares
their role specialization and structure-grounded retrieval, but differs
on two axes. First, retrieval operates on a typed, empirical
\emph{dataset co-usage} graph with a heterogeneous GNN ranker rather
than per-query literature retrieval or symbolic ontology traversal, so
every generated hypothesis is pinned to two named NASA products and is
actionable by construction. Second, prior pipelines typically fix one
LLM per role and ---in the agentic case--- equip that LLM 
with execution tools;  we instead cross agent identity in a
$2{\times}2{\times}2$ factorial (\cref{sec:factorial}) and report the
resulting variance decomposition, which is particularly consequential
when the reward itself is LLM-defined, as in
\citet{wang2024scimon} and \citet{agarwal2025autodiscovery}.

\paragraph{LLM-as-judge and evaluation reliability.}
The reliability of LLM-based evaluation has been studied extensively.
\citet{zheng2023mtbench} introduce the LLM-as-judge formulation and
document position, verbosity, and self-enhancement biases.
\citet{kenton2024oversight} analyze scalable-oversight protocols in
which weak LLM judges adjudicate strong LLM agents, identifying
positional and order effects across nine reasoning asymmetries.
\citet{ye2024calm} catalog twelve bias types and automate their
quantification via CALM; \citet{wataoka2024selfpreference} isolate
self-preference bias and link it to perplexity, predicting that a
judge will prefer outputs whose distribution is familiar to it. Most
recently, \citet{yang2025rbd} propose a plug-in Reasoning-based Bias
Detector for post-hoc correction. Our factorial contributes an
\emph{empirical} variance decomposition in the scientific-hypothesis
evaluation regime: judge identity accounts for roughly $25\%$ of
importance-score variance, dwarfing filter or generator effects
(${<}2\%$), and the blind-versus-contextual contrast dominates
tractability variance (${\sim}16\%$). This is precisely the failure
mode anticipated by \citet{wataoka2024selfpreference} and
\citet{ye2024calm}, and the one \citet{yang2025rbd} aims to mitigate,
now quantified in the hypothesis-evaluation setting for the first time.

\paragraph{Machine learning for Earth observation.}
\citet{rolf2024satellite} argue that satellite data is a distinct ML
modality and call for dedicated infrastructure. The dominant response
has been to scale foundation models over individual imagery products
\citep{stewart2023ssl4eol, guo2024skysense, wu2025skysensepp},
surveyed in \citet{xiao2024rssurvey}. At the community level,
\citet{zhu2025earthfm} articulate eleven desiderata for ideal Earth
foundation models and explicitly flag cross-dataset knowledge synthesis
as an open gap. Our work targets precisely that complementary
discovery layer: given hundreds of heterogeneous NASA products, which
\emph{pairings} are worth a researcher's time? We address this question here.

\section{Knowledge Graph and GNN Retrieval}
\label{sec:kg-gnn}

\paragraph{Knowledge graph.} We build on the publicly released NASA
Earth Observation Knowledge Graph (EO-KG) \citep{nasa_eokg_2024}:
a single GraphML resource with 150{,}351 nodes and 436{,}203 edges,
spanning seven node types (\texttt{Publication}, \texttt{Dataset},
\texttt{ScienceKeyword}, \texttt{Instrument}, \texttt{Platform},
\texttt{Project}, \texttt{DataCenter}) and nine typed relations
covering citation, dataset usage, metadata membership, and a
science-keyword subcategory hierarchy. The ranking universe consists of the 1{,}475 datasets that participate in at least one training-period co-usage pair (i.e., are co-cited with another dataset by at least one paper with year~$\le 2022$). This is by construction the natural evaluation pool for our link-prediction task and is also the candidate pool used for negative sampling during training. Per-type counts and the full relation breakdown are in \cref{app:kg-stats}.

\paragraph{Deriving co-usage supervision.} The raw KG contains no
direct dataset--dataset edge; we derive co-usage from
\texttt{Publication$\to$Dataset} edges. For every publication~$p$
with dataset set $D_p = \{d_{p,1}, \dots, d_{p,k}\}$, $k \ge 2$, and
every unordered pair $(d_i, d_j) \in D_p$, we emit a co-usage
observation tagged with the publication year. Pairs are labeled
binary (co-used vs.\ not); multiplicity is retained as a weight only
for auxiliary analyses. We then split pairs temporally by the publication year of each paper that observes them: train ($\le 2022$), val ($=2023$), test
($=2024$), giving 13{,}529 / 6{,}119 / 6{,}319 pairs. A pair
co-cited in papers from multiple years enters multiple splits; we
report \emph{test pairs unseen in train} (3{,}944) as the strict
held-out subset used for evaluation. Within test
we carve two harder subpools: \emph{cold-start}
(2{,}284 pairs, at least one endpoint with no training co-usage) and
\emph{cross-DAAC} (1{,}480 pairs, endpoints in distinct archives).
Full statistics are reported in \cref{tab:cousage_stats}.

\paragraph{Heterogeneous GNN ranker.}
Let $\mathcal{G}$ be the training graph induced by five surviving
node types (\texttt{Dataset}, \texttt{ScienceKeyword},
\texttt{Instrument}, \texttt{Platform}, \texttt{Project}) and six
typed relations: \texttt{co\_usage} between datasets, plus
\texttt{has\_platform}, \texttt{has\_keyword},
\texttt{has\_instrument}, \texttt{of\_project}, and
\texttt{has\_subcategory}. Each \texttt{Dataset} node is initialized
with a 768-dim SPECTER2 abstract embedding
\citep{singh2023specter2, cohan2020specter}; non-dataset types
receive Xavier-initialized learnable embeddings of the same
dimension. Publication nodes are removed from $\mathcal{G}$:
they served only to derive co-usage supervision, and retaining them
would leak test pairs through shared neighbors.

We learn $d{=}128$ dimensional node representations with a two-layer
heterogeneous \mbox{GraphSAGE} \citep{zhang2018seal, wang2024ncn}.
Within each layer $\ell$, for every typed relation $r$ and every
target node $v$ reachable through $r$, we form a per-relation
message
\begin{equation}
    \mathbf{m}_v^{(\ell+1,\,r)}
    \;=\;
    \mathbf{W}_r^{(\ell)} \,
    \mathrm{MEAN}_{u \in \mathcal{N}_r(v)}\, \mathbf{h}_u^{(\ell)},
\end{equation}
where $\mathcal{N}_r(v)$ are the neighbors of $v$ under~$r$.
Messages incoming to $v$ are summed across all incident relations
and combined with the self-state through a SAGE residual,
\begin{equation}
    \mathbf{h}_v^{(\ell+1)}
    \;=\;
    \sigma\!\Big(
        \mathbf{W}_{\mathrm{self}}^{(\ell)} \mathbf{h}_v^{(\ell)}
        + \!\!\sum_{r \in \mathcal{R}_v}\! \mathbf{m}_v^{(\ell+1,\,r)}
    \Big),
\end{equation}
where $\mathcal{R}_v$ is the set of relations incident on $v$,
$\sigma {=} \mathrm{ReLU}$, and dropout $0.2$ is applied to
$\mathbf{h}_v^{(\ell+1)}$. Per-relation weights
$\mathbf{W}_r^{(\ell)}$ are \emph{not} shared across relations; this
is what distinguishes the heterogeneous variant from a homogeneous
SAGE applied to the union graph and what allows the model to respect
the feature-heterophilic structure inherent to typed graphs
\citep{zhu2024heterophily}. We use $L{=}2$ layers; the
\texttt{(Dataset, co\_usage, Dataset)} relation participates in
message passing on \emph{training pairs only}.

The final dataset embeddings $\mathbf{h}_i \!=\! \mathbf{h}_i^{(L)}$
are passed through one of two scorers,
\begin{align}
    z_{\mathrm{dot}}(i, j) &= \mathbf{h}_i^{\!\top}\, \mathbf{h}_j, \\
    z_{\mathrm{mlp}}(i, j) &= \operatorname{MLP}\!\big([\mathbf{h}_i \,\|\, \mathbf{h}_j]\big),
\end{align}
with the MLP being a two-layer network with hidden width $d$ and
\textsc{ReLU} activation. Both produce a logit; predicted
co-usage probability is $\sigma(z(i,j))$. We report both heads
because, as we show below, they form a Pareto pair across the
evaluation pools.

\paragraph{Training objective.} Stage~2 is trained by binary
cross-entropy with logits. Each minibatch~$\mathcal{B}$ samples one
training positive $(s, t^{+})$ together with one negative
$(s, t^{-})$, and minimizes
{\small
\begin{equation}
\mathcal{L}(\mathcal{B})
=
-\sum_{(s,t^{+},t^{-}) \in \mathcal{B}}
\Big[
\log \sigma(z(s,t^{+}))
+
\log\big(1 - \sigma(z(s,t^{-}))\big)
\Big].
\end{equation}
}
Negatives are drawn from a degree-biased proposal,
$\Pr(t^{-} \!=\! n) \!\propto\! \deg(n)^{0.75}$
(matching the \textsc{word2vec} convention of upweighting hub
candidates so that the ranker is forced to discriminate against
plausible distractors), with rejection of any~$n$ that is itself a
training positive of~$s$. Optimization uses Adam at learning rate
$10^{-3}$ for up to 300 epochs, with early stopping on validation MRR
(patience 30). 

\paragraph{Evaluation protocol.}
We evaluate as a 1{:}100 ranking task: each held-out positive
$(a, b^{+})$ is scored against 100 degree-matched negatives
$\{b^{-}_k\}_{k=1}^{100}$, and we report Hits@$k$, mean reciprocal
rank (MRR), AUC, and average precision (AP) on each of the three
pools defined above. Hits@$k$ is the fraction of held-out positives
ranked in the top~$k$ over the 101-element candidate list; MRR is
the mean of $1/\mathrm{rank}(b^{+})$. 

\paragraph{Baselines.}
We compare against six baselines spanning structure-only and
content-only signals.
\textbf{Popularity} ranks candidates by global degree:
$z(i, j) = \deg(j)$, ignoring~$i$ entirely.
\textbf{Common Neighbors} and \textbf{Adamic--Adar} score the
candidate by overlap of co-usage neighborhoods,
$z(i, j) = \sum_{u \in \mathcal{N}(i) \cap \mathcal{N}(j)} \!w(u)$,
with $w \!\equiv\! 1$ for the former and
$w(u) \!=\! 1/\log|\mathcal{N}(u)|$ for the latter.
\textbf{MF-SVD} factorizes the binary publication$\times$dataset
co-citation matrix by truncated SVD ($k=128$) and scores pairs by
inner product of the resulting dataset rows.
\textbf{SPECTER2} \citep{singh2023specter2} and
\textbf{BGE-base} \citep{xiao2024cpack} are dense-retrieval baselines that score pairs by
cosine similarity of the corresponding dataset abstract embeddings,
giving us a content-only ceiling to compare the GNN against.

\begin{table}[t]
\centering
\small
\setlength{\tabcolsep}{4pt}
\caption{Hits@10 on three test-set pools (pre-2023 train, 2024 test) under 1{:}100 ranking, where random performance is Hits@10 $\approx 0.10$. \textbf{Bold}: best per column. Content baselines fail cross-DAAC; structural baselines fail cold-start; the heterogeneous GNN is the first method competitive on all three pools, reaching $4.7{\times}$ random on the full pool and $5.2{\times}$ random on cold-start. The two scorer heads form a Pareto pair: dot for cold-start (0.519), MLP for cross-DAAC (0.266). All publication-derived edges are restricted to year~$\le 2022$.}
\label{tab:gnn_baselines}
\begin{tabular}{lccc}
\toprule
Method & All & Cold-start & Cross-DAAC \\
\midrule
Popularity & 0.140 & 0.064 & 0.130 \\
MF-SVD & 0.192 & 0.274 & 0.125 \\
Common Neighbors & 0.349 & 0.211 & 0.228 \\
Adamic-Adar & 0.355 & 0.225 & 0.235 \\
BGE-base & 0.394 & 0.484 & 0.103 \\
SPECTER2 & 0.413 & 0.477 & 0.127 \\
GNN-Homo & 0.448 & 0.473 & 0.146 \\
\textbf{GNN-Hetero (dot)} & \textbf{0.468} & \textbf{0.519} & 0.175 \\
\textbf{GNN-Hetero (MLP)} & 0.467 & 0.389 & \textbf{0.266} \\
\bottomrule
\end{tabular}
\end{table}

\paragraph{Results.}
\Cref{tab:gnn_baselines} reports Hits@10 across the three pools, and
the cross-pool pattern is sharp. \textbf{Popularity} is essentially a
prior on dataset frequency and accordingly performs near floor on
\textbf{all} (0.140) and \textbf{cold-start} (0.064). The two
co-usage-graph heuristics, \textbf{Common Neighbors} (0.349) and
\textbf{Adamic--Adar} (0.355), more than double Popularity on
\textbf{all}, but their reliance on observed neighbors is fatal on
\textbf{cold-start} (0.211, 0.225), where at least one endpoint has
no training neighborhood at all. \textbf{MF-SVD} shows the opposite
asymmetry: its 128-dim factorization happens to generalize on
\textbf{cold-start} (0.274) but loses on \textbf{cross-DAAC} (0.125),
where bipartite co-citation structure does not constrain inter-archive
pairs. Content baselines invert this picture again:
\textbf{SPECTER2} and \textbf{BGE-base} dominate \textbf{cold-start}
(0.477 and 0.484), confirming that abstract semantics is the strongest
single signal when no structural cue is available, but they collapse
on \textbf{cross-DAAC} (0.127, 0.103). The intuition is that
abstracts of cross-archive datasets share substantial Earth-science
boilerplate, so cosine similarity ceases to discriminate plausible
joint analyses from coincidental textual overlap.

The two GNN variants combine these signals. \textbf{GNN-Homo}
matches the content baselines on \textbf{cold-start} (0.473) and
exceeds them on \textbf{all} (0.448), but gains little on
\textbf{cross-DAAC} (0.146). \textbf{GNN-Hetero}, which carries
type-specific aggregators across the metadata relations, is the first
method competitive on all three pools simultaneously: it improves
\textbf{cold-start} to 0.519 and lifts \textbf{cross-DAAC} to 0.175.
The two scorer heads form a clear Pareto pair: the dot-product head
peaks on \textbf{cold-start} (0.519) while the MLP head peaks on
\textbf{cross-DAAC} (0.266)---a substantial $0.091$ absolute gain
over its dot-product counterpart, attributable to the MLP's ability
to learn nonlinear combinations of the two embeddings that compensate
for the type-confounded similarity between cross-archive datasets.
We deploy the dot-product head as the retrieval feed to Agent~1
(selected by validation MRR over the \textbf{all} pool) and release
both embedding sets with the code; the per-pool, per-metric breakdown
for all configurations and the 11-lever ablation that selected this
configuration are in \cref{app:gnn-baselines-full} and
\cref{app:gnn-ablation}.

\section{Three-Agent LLM Pipeline}
\label{sec:pipeline}

The 200 predicted-novel pairs surfaced by the GNN
(\cref{sec:kg-gnn}) are the input to a three-agent LLM cascade.
Each agent has a distinct role: scoring, generation, and judgment. Each agent is independently instantiated with one of two backbones
(GPT-5.2 or Claude Sonnet~4.6). All three agents return strictly
JSON-formatted output with a fixed schema, queried at temperature~$0$
to remove sampling variance from the experiment. The full system
prompts, user templates, and JSON schemas for all four roles
(Agent~1, Agent~2, Agent~3 blind, Agent~3 contextual) are reproduced
verbatim in \cref{app:prompts}.

\paragraph{Agent 1 — pair-level filter.}
Agent~1 receives a candidate pair $(d_i, d_j)$ together with the two
datasets' short names, long names, and abstracts (truncated at
1{,}200 characters), and returns two integer scores on a $1$--$5$
scale: \emph{plausibility} (could a research team reasonably combine
these datasets in a single study?) and \emph{novelty} (how
non-obvious is the combination?), plus a two-to-three sentence
rationale. Scoring is deliberately decoupled from generation: Agent~1
sees no hypothesis text, so its rating reflects the pair itself, not
how well a generator happens to motivate it. To anchor the scoring
distribution and provide a positive control, Agent~1 is run not only
on the 200 predicted-novel pairs (stratum~A) but also on three
additional strata of equal size containing held-out real co-usages
sampled from 2024 publications, giving 800~scored pairs per Agent~1
backbone. Because the held-out pool is what the GNN was \emph{not}
trained on, comparing the plausibility distribution of stratum~A
against strata B--D is a direct test of whether predicted-novel
pairs are scientifically coherent at the level of literature co-usages
(\cref{sec:results}).
For each Agent~1 backbone we select 40 pairs to forward to the
generator, drawn exclusively from stratum~A. We use a tiered rule
rather than ranking by a single composite score, because plausibility
and novelty answer different questions and naive scalarization would
let either dimension dominate. Tier~1 takes pairs with maximal
plausibility (${=}5$) and at least moderate novelty ($\ge 3$);
tier~2 takes pairs with high plausibility ($=4$) and high novelty
($\ge 4$); tier~3 takes pairs with maximal plausibility ($=5$) and
modest novelty ($=2$), which capture obvious-yet-strong pairings as
a sanity floor. Within each tier, pairs are ordered by a weighted composite score,
giving 55\% weight to plausibility and 45\% to novelty, and the top
40 are taken in order Tier~1 $\to$ Tier~2 $\to$ Tier~3, with a
fallback to all $\text{plaus} \ge 4$ pairs ranked by the same score
if the three tiers yield fewer than 40.

\paragraph{Agent 2 — hypothesis generator.}
Agent~2 receives one selected pair $(d_i, d_j)$ together with their
short names, long names, and abstracts, and returns a structured
research hypothesis with six fields: a one-sentence
\emph{research question}, a one-to-two sentence testable
\emph{hypothesis}, an \emph{analysis method} describing how the two
datasets jointly test it, an \emph{expected finding} that would
constitute support, a one-to-two sentence statement of
\emph{scientific importance}, and a one-to-three word \emph{domain}
tag. Critically, Agent~2 does not see Agent~1's plausibility or
novelty scores; this prevents the generator from regressing toward
its own filter's preferences and keeps generation a reflection of
what the backbone independently considers worth proposing.

\paragraph{Agent 3 — judge under blind and contextual conditions.}
Agent~3 evaluates each generated hypothesis on three integer scales:
\emph{importance} (does it meaningfully advance the field?),
\emph{tractability} (is it ready to execute with current methods?),
and \emph{novelty} (does it open a new line of inquiry?). Each
hypothesis is judged twice---once \emph{blind} (Agent~3 sees only the
six hypothesis fields, with no information about which datasets
underlie them) and once \emph{contextual} (Agent~3 sees the
hypothesis fields plus the underlying pair's short names, long names,
and abstracts). The blind/contextual contrast isolates how much of
the judgment depends on the hypothesis text alone versus on
inspectable evidence about the datasets. We will show in
\cref{sec:results} that this contrast is the single largest source
of variance in tractability scores, dwarfing both filter and
generator effects.


\section{Factorial Experiment}
\label{sec:factorial}

To diagnose how much each agent contributes to the quality scores
the pipeline reports, we run the three-agent cascade
(\cref{sec:pipeline}) under all eight assignments of backbone to role
in a $2{\times}2{\times}2$ factorial design. The three factors are
the identities of the filter (Agent~1), the generator (Agent~2),
and the judge (Agent~3); each is independently set to either GPT-5.2
or Claude Sonnet 4.6. A fourth nested factor, judging
\emph{condition}, varies whether the judge sees the underlying
dataset metadata (\textsc{contextual}) or only the hypothesis text
(\textsc{blind}). All other components of the pipeline including the GNN
ranker, the four-stratum input pool fed to Agent~1, the
tiered top-40 selection, prompt templates, JSON schemas, and the
temperature-$0$ decoding are held constant across cells, so any
observed differences are attributable to backbone identity or to
the blind/contextual contrast.

\paragraph{Cells and counts.}
The factorial yields four hypothesis sets indexed by $(a_1, a_2)$,
each containing 40 hypotheses, for a total of 160 distinct
hypotheses. Each hypothesis is then scored by both judge backbones
$a_3 \in \{\text{GPT}, \text{Claude}\}$ under both judging
conditions, producing $160 \times 4 = 640$ validator judgments.
Each judgment returns three integer scores on a $1$--$5$ scale
(importance, tractability, novelty) plus a free-text rationale.
Agent~1 itself is also a factorial cell with 800 inputs (4 strata
of 200 pairs), giving $2 \times 800 = 1{,}600$ Agent~1 ratings used
both to feed the top-40 selection and to characterize how predicted-
novel pairs (stratum~A) compare against held-out real co-usages
(stratum~B) and two control strata
(\cref{sec:results}).

\paragraph{Why factorial.}
The dominant practice in LLM-as-scientist evaluations is to fix one
backbone in each role and report the resulting scores as if they
were properties of the hypotheses being judged. The factorial
design instead lets us \emph{decompose} score variance across the
four nominal factors (filter, generator, judge, condition) and
their two-way interactions using a standard one-way ANOVA per axis,
reporting effect sizes as $\eta^2$. This decomposition asks a
direct question: when an importance score moves from~3 to~4, how
much of that movement is signal about the hypothesis, and how much
is the contingent identity of the backbones playing each role? We
will show that for some axes the answer is uncomfortable.

\section{Results}
\label{sec:results}

\begin{figure*}[t]
\centering
\includegraphics[width=0.92\linewidth]{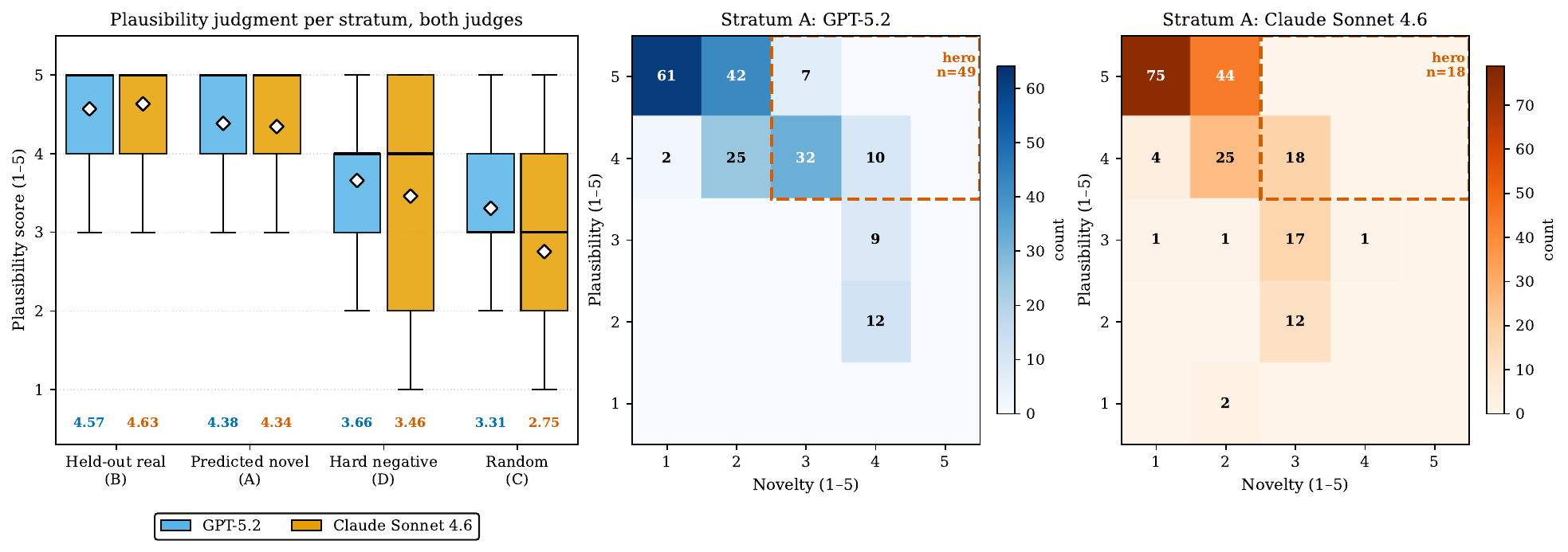}
\caption{Per-judge $(\text{plausibility}, \text{novelty})$ score
grid for stratum~A (GNN-predicted novel pairs). Cells report counts
out of 200; rows are plausibility ($5$ at top) and columns are
novelty ($1$ at left). The dashed red rectangle delimits the
\emph{hero zone} (plausibility $\ge 4$ and novelty $\ge 3$) targeted
by the tiered top-40 selection of \cref{sec:pipeline}. GPT-5.2
populates the hero zone more densely than Claude Sonnet 4.6 ($n=49$
vs.\ $n=18$), reflecting Claude's tighter novelty distribution
rather than a difference in plausibility ranking; both judges
concentrate stratum-A pairs in the high-plausibility rows
($\ge 4$).}
\label{fig:stage4_judgment}
\end{figure*}

\paragraph{Stage 4: predicted-novel pairs are nearly as plausible as
held-out real co-usages.}
\Cref{tab:stage4_strata} reports the Agent-1 stratum-level results
and \cref{fig:stage4_judgment} visualizes the
$(\text{plausibility}, \text{novelty})$ joint distribution for
stratum~A under each judge. Both judges order the four strata
identically: held-out real 2024 co-usages (B) on top, GNN-predicted
novel pairs (A) a hair below, then same-DAAC random ``hard
negatives'' (D), and fully random pairs (C) at the bottom.
Critically, the gap between A and B on plausibility is tiny---0.19
for GPT (4.38 vs.\ 4.57) and 0.29 for Claude (4.34 vs.\ 4.63)---
while the gap between A and the random control C is more than five
times larger ($\ge$1.07 for GPT, $\ge$1.59 for Claude). The GNN
therefore surfaces predicted-novel pairs that LLM judges find
essentially indistinguishable in scientific plausibility from pairs
that were actually co-used in 2024 publications. The novelty axis,
by contrast, fails to separate A from B: judges rate the
predicted-novel pairs as no more novel than held-out real co-usages,
a pattern we return to in \cref{sec:discussion} as a caution about
LLM-as-novelty-judge. The joint distribution in
\cref{fig:stage4_judgment} additionally exposes a calibration shift
between judges: 49 of 200 stratum-A pairs land in GPT's hero zone
(plausibility $\ge 4$ and novelty $\ge 3$), versus only 18 in
Claude's, reflecting a tighter Claude novelty distribution
(\cref{tab:stage4_strata}) rather than a difference in plausibility
ranking. This calibration mismatch is the practical reason we run
the tiered top-40 selection independently per Agent-1 backbone
rather than pooling.

\begin{table}[t]
\centering
\caption{Stage-4 pair-level judgments. Each of 800 pairs (200 per
stratum) was scored by both backbones on plausibility and novelty
($1$--$5$). \textbf{B} = held-out real co-usages from 2024
publications; \textbf{A} = GNN-predicted novel pairs;
\textbf{D} = same-DAAC random ``hard negatives''; \textbf{C} = fully
random pairs. Both judges order the strata identically
(B $\!\geq\!$ A $>$ D $>$ C on plausibility); the A--B gap is small ($\le 0.29$) while the A--C gap is more than five times larger.}
\label{tab:stage4_strata}
\small
\setlength{\tabcolsep}{4pt}
\begin{tabular}{lcccc}
\toprule
\multirow{2}{*}{Stratum} & \multicolumn{2}{c}{Plausibility} & \multicolumn{2}{c}{Novelty} \\
\cmidrule(lr){2-3}\cmidrule(lr){4-5}
& GPT-5.2 & Claude 4.6 & GPT-5.2 & Claude 4.6 \\
\midrule
B & 4.57 & 4.63 & 2.35 & 1.86 \\
A & 4.38 & 4.34 & 2.19 & 1.84 \\
D & 3.66 & 3.46 & 3.17 & 2.48 \\
C & 3.31 & 2.75 & 3.52 & 2.75 \\
\bottomrule
\end{tabular}
\end{table}

\begin{figure}[t]
\centering
\includegraphics[width=\linewidth]{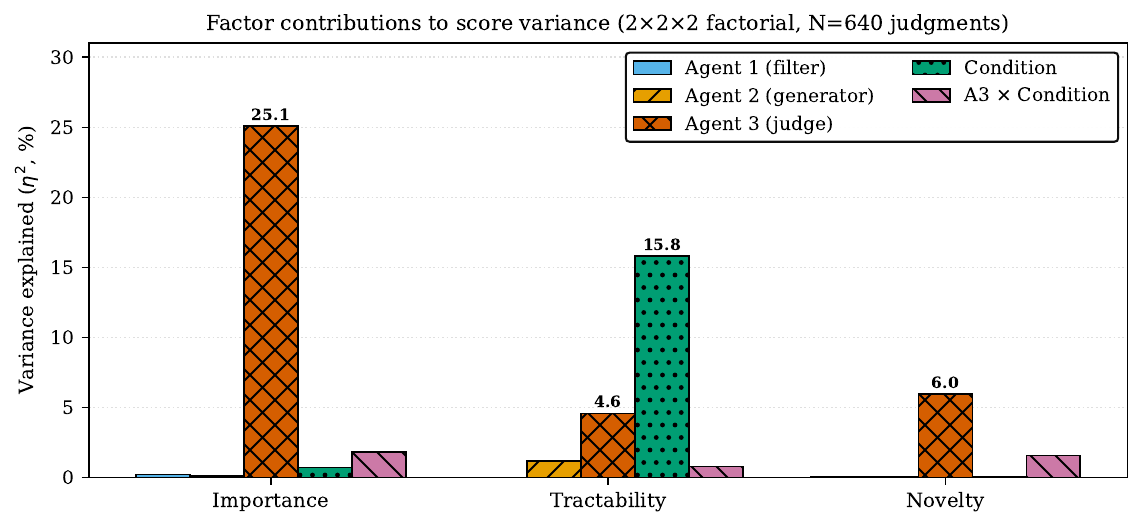}
\caption{Variance decomposition of validator scores by factor, per
axis ($\eta^2$ from one-way ANOVA, $N=640$ judgments). Judge
identity dominates importance variance; the blind/contextual
contrast dominates tractability variance; no factor explains more
than $\sim$10\% of novelty variance. Filter and generator identity
are negligible on every axis.}
\label{fig:factorial_variance}
\end{figure}

\paragraph{Stage 5: variance decomposition.}
\Cref{fig:factorial_variance} reports the $\eta^2$ contribution of
each factor to score variance, by axis. Three patterns dominate.
\textbf{(i) Importance is a judge-identity question.} Judge identity
explains $25.1\%$ of importance variance, while filter and generator
identity together explain $0.34\%$. In absolute terms the GPT judge
rates importance roughly $0.6$~points higher than the Claude judge
in nearly every cell, regardless of who filtered or generated the
hypothesis. \textbf{(ii) Tractability is a context question.} The
blind/contextual contrast explains $15.8\%$ of tractability variance
versus $4.6\%$ for judge identity; showing the dataset metadata
reliably lowers tractability scores by about half a point
(blind mean $3.94$, contextual mean $3.36$, averaged over the eight
$(a_1,a_2,a_3)$ cells), suggesting that hypothesis text alone reads
as more executable than it does once the actual measurement
products are inspectable. \textbf{(iii) Novelty is judge-dominated
but small.} Judge identity is again the largest single factor
($6.0\%$), but no factor explains more than $\sim$10\% of variance
on novelty---consistent with the Stage-4 finding that LLM judges
do not strongly discriminate novelty across pair types. Filter and
generator identity contribute essentially zero variance on every
axis ($\eta^2 < 0.013$ in all six cells), which has a useful
corollary: Agent~1 and Agent~2 affect \emph{which} hypotheses get
scored, but their backbone identity does not bias the scores those
hypotheses receive.

\paragraph{Inter-rater agreement: judges line up on tractability,
diverge on importance.}
Inter-rater agreement between GPT-5.2 and Claude Sonnet~4.6 on
identical hypotheses (full table in \cref{app:factorial}) is highest
on tractability (Pearson $r=0.63$ blind, quadratic-weighted
$\kappa_q=0.47$), moderate on novelty ($r=0.49$, $\kappa_q=0.36$),
and weakest on importance ($r=0.35$, $\kappa_q=0.14$). The ordering
is the opposite of what one might expect---importance is arguably
the most subjective of the three rubrics---but it is exactly what
\citet{wataoka2024selfpreference} predict for an LLM-as-judge setup:
importance is the axis most coupled to a backbone's training-time
priors about what counts as ``meaningful advance,'' and consequently
the axis on which two backbones with different pretraining mixtures
disagree most. Showing dataset context improves agreement on
\emph{all} three axes, with the largest gain on importance
($\kappa_q$: $0.14 \to 0.24$, exact match: $37\% \to 63\%$),
supporting the hypothesis that anchoring judges to inspectable
evidence narrows the prior-driven disagreement.

\paragraph{Hypothesis ranking is stable across configurations.}
The variance decomposition above is about the \emph{absolute}
scores, but for picking which hypotheses to actually pursue, what
matters is the \emph{ranking}---which hypothesis scores higher than
which. That ranking turns out to be much more stable than the
absolute scores. Across the four judge configurations (each pairing
of judge backbone and blind/contextual condition), 159 of 160
hypotheses change by at most 2 points on every axis, with an
average change of just $1$ point. So even though one judge gives
systematically higher scores than the other, the order in which
they rank the hypotheses is nearly the same. A researcher who picks
the top $25\%$ by average score will get almost the same set no
matter which single judge is used. The practical takeaway:
single-judge scores are not portable in absolute terms, but the
rankings they produce are reliable.

\paragraph{Flagship hypotheses.}
A qualitative inspection of the highest-scoring outputs is deferred
to \cref{app:flagship}, which lists five flagship hypotheses with
the highest combined importance + tractability and a maximum cross-
judge disagreement of one on every axis. The five span ecohydrology,
glaciology, aerosol--cloud interactions, vegetation phenology, and
methodological bias correction in cloud retrieval; all five are
pinned to two named NASA products and are executable today with
public archives---the deployment property the pipeline was designed
to preserve.

\section{Discussion and Limitations}
\label{sec:discussion}

\paragraph{Scope: ideation, not execution.} Our three agents
operate text-only by design: filter and judge are evaluative roles
where tool use buys little, and generator is an ideation role
whose output is intended to seed human-led investigation rather
than autonomously execute it. This scope choice has a second
benefit specific to our methodology: the variance decomposition of
\cref{sec:results} is interpretable precisely because every
backbone receives identical inputs at temperature $0$ with strict-
JSON decoding. Adding tool access would inject retrieval and
execution stochasticity that would confound the agent-identity
signal we set out to measure. We view text-only ideation and
tool-equipped execution as complementary layers of a future
end-to-end discovery system, with our pipeline supplying the
candidate hypotheses an executor would attempt. The most important limitation of this work is that the 160
hypotheses we report are LLM-judged, not human-verified. A small
domain-expert study---even on the five flagship hypotheses---would
sharpen the claim that the pipeline produces actionable research
seeds rather than merely well-formed text. Three further limitations
deserve mention. First, the GNN treats co-usage as a binary signal
and discards multiplicity; an edge-weight-aware variant might
sharpen the ranker but is not the bottleneck identified by our
ablation (\cref{app:gnn-ablation}). Second, abstracts are truncated
at 1{,}200 characters before being shown to any agent, which can
clip relevant detail for long-abstract products and may
under-represent some pairings. Third, our factorial covers two
backbones, two judging conditions, and three rubric axes; it does
not control for prompt phrasing, position bias within the user
template, or temperature, all of which prior work has shown to
move LLM-judge outputs \citep{zheng2023mtbench, ye2024calm}.
Quantifying their interaction with the variance budget reported
here is a natural next step.

\section{Conclusion}
\label{sec:conclusion}

Hypothesis generation in observational sciences is a different
problem from generation in literature-driven ones: the unit of
discovery is a specific pair of measurement products, not a
sentence. We addressed this by combining a heterogeneous GNN
ranker over the NASA Earth-Observation Knowledge Graph with a
three-agent LLM cascade, producing 160 structured, dataset-grounded
research hypotheses spanning many Earth-science domains. Predicted-
novel pairs are rated essentially as plausible as held-out real
co-usages, and a $2{\times}2{\times}2$ factorial over agent identity
shows that hypothesis rankings are stable across LLM backbones while
absolute scores are not---an empirical bound on what single-judge
LLM-as-scientist evaluations can claim. We release the trained GNN,
the 160 hypotheses, the 640 validator judgments, and the full
pipeline as a reproducible substrate for both the EO community and
the broader study of LLM-driven scientific discovery.


\section*{Impact Statement}

This paper presents a system that suggests candidate research directions
in Earth science; generated hypotheses are not scientific claims but
seeds for human investigation, and each must be verified with domain
experts and actual data analysis before being treated as a finding.
We see no societal consequences requiring specific discussion beyond
standard concerns about the use of LLM outputs in scientific contexts.

\section*{LLM Usage Disclosure}

GPT-5.2 and Claude Sonnet 4.6 were used \emph{as pipeline components}:
they are the Agents~1, 2, and 3 described in \cref{sec:pipeline}, and
their outputs are the experimental data reported in \cref{sec:results}.
In addition, Claude was used as a writing collaborator during drafting of
this manuscript; all final wording and technical claims were verified by
the authors.

\bibliography{main}
\bibliographystyle{icml2026}

\appendix
\onecolumn
\section{Knowledge Graph Characterization}
\label{app:kg-stats}

\begin{figure}[t]
\centering
\includegraphics[width=0.85\linewidth]{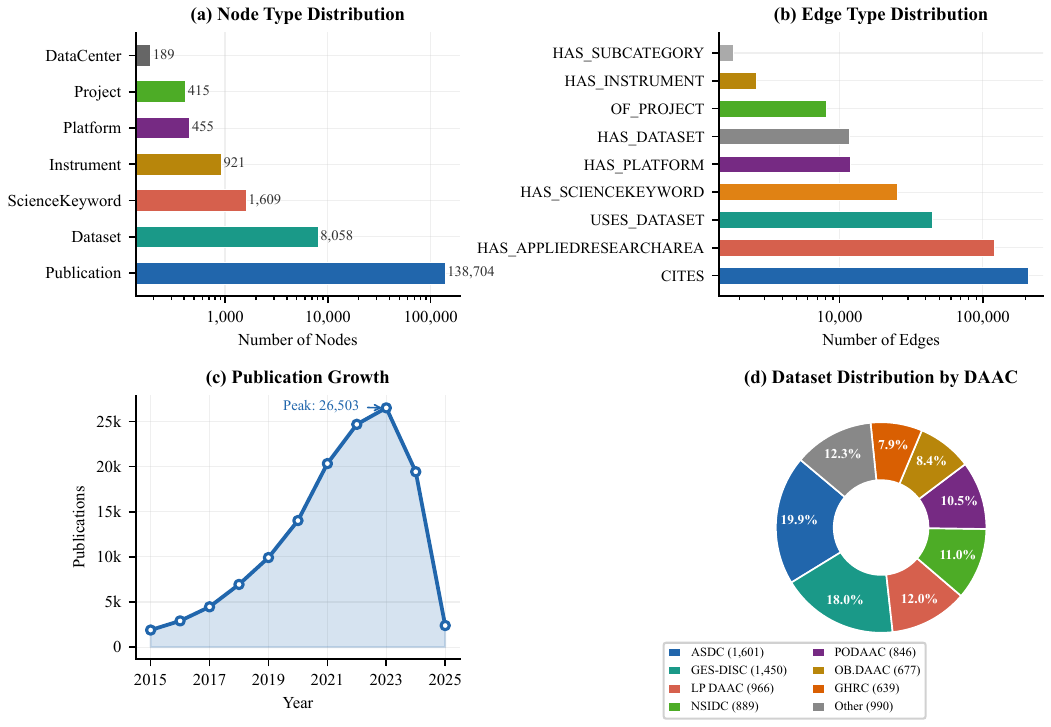}
\caption{Node-type and edge-type distributions in the NASA EO-KG.
Publications dominate the node count; citations and research-area
edges dominate the edge count.}
\label{fig:graph_overview}
\end{figure}

\begin{figure}[t]
\centering
\includegraphics[width=0.85\linewidth]{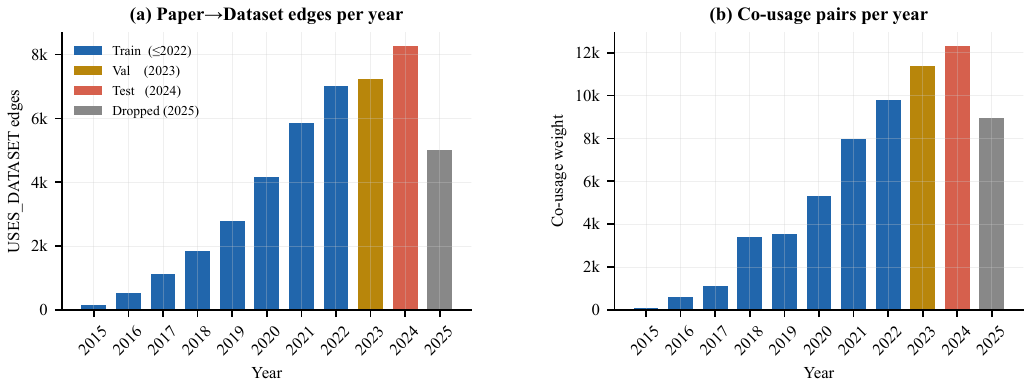}
\caption{Temporal split of the NASA EO-KG by publication year, colored
by assigned split (blue: train, $\le 2022$; gold: val, $2023$; orange:
test, $2024$; grey: $2025$, dropped for incompleteness).
\textbf{(a)}~Number of \texttt{USES\_DATASET} (Paper$\to$Dataset)
edges emitted per year; this is the raw scaffolding from which
co-usage is derived.
\textbf{(b)}~Total co-usage weight per year, i.e.\ the number of
unordered dataset pairs contributed by each year's papers (a paper
citing $k$ datasets contributes $\binom{k}{2}$ to its year).
Each co-usage pair is assigned to every split corresponding to a
year in which a paper co-citing both datasets appeared; pairs cited
across multiple years thus enter more than one split. The split
sizes are 13{,}529 training, 6{,}119 validation, and 6{,}319 test
pairs, with 3{,}944 test pairs unseen in train serving as the
strictly held-out evaluation subset.}
\label{fig:temporal_split}
\end{figure}

\begin{figure}[t]
\centering
\includegraphics[width=0.85\linewidth]{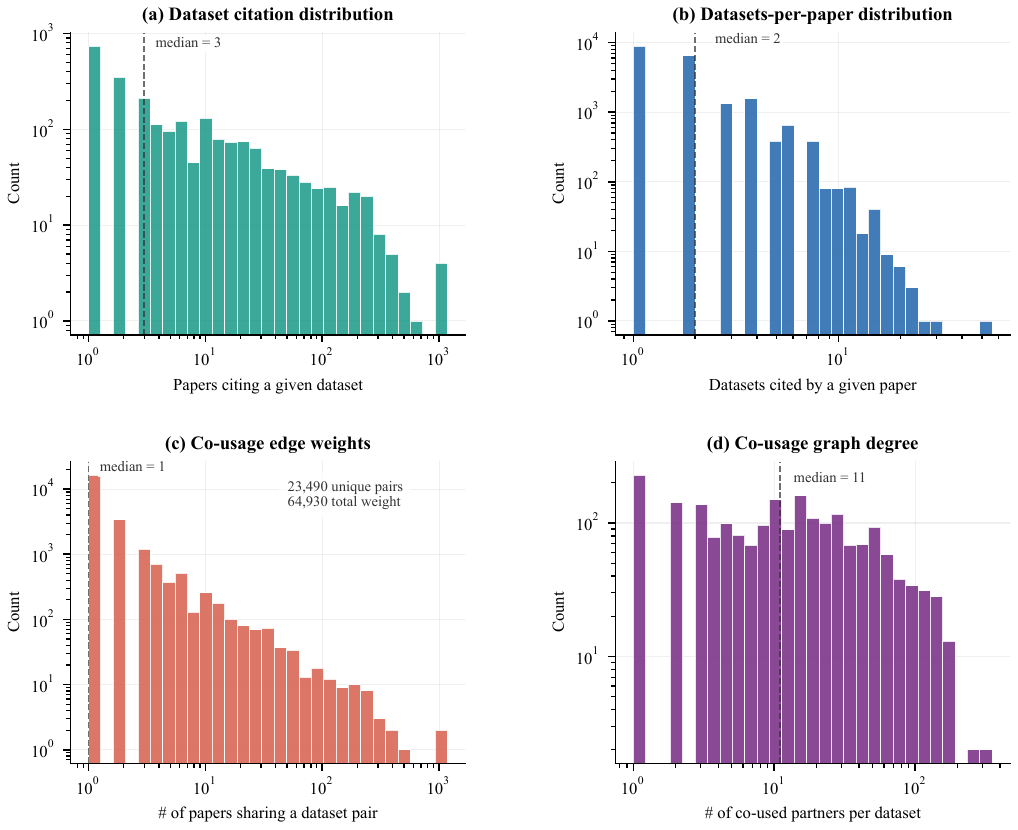}
\caption{Usage and co-usage distributions in the NASA EO-KG, all on
log--log axes.
\textbf{(a)}~Number of papers citing a given dataset
(median 3, max \num{1051}).
\textbf{(b)}~Number of datasets cited by a given paper
(median 2, max 50).
\textbf{(c)}~Co-usage \emph{edge weight}: for each unordered pair of
datasets that have ever been jointly cited, the number of distinct
papers that cite both (max pair weight \num{1051}; \num{23490} unique
pairs sum to total weight \num{64930}).
\textbf{(d)}~Co-usage \emph{graph degree}: for each dataset, the number
of distinct other datasets it has ever been co-cited with
(median 11, max 314). All four distributions are heavy-tailed, a
small number of flagship datasets (MODIS, MERRA-2, SMAP variants) and
flagship pairings dominate the graph.}
\label{fig:usage_overview}
\end{figure}

This appendix reports the full statistics of the NASA EO Knowledge
Graph \citep{nasa_eokg_2024} summarized in \cref{sec:kg-gnn}.
\Cref{tab:graph_stats} lists global graph properties,
\cref{tab:node_types} the seven node types,
\cref{tab:edge_types} the nine relation types, and
\cref{tab:daac} the dataset distribution across NASA's Distributed
Active Archive Centers, and \cref{tab:cousage_stats} the derived
co-usage statistics and temporal split sizes. \Cref{fig:graph_overview} visualizes node-
and edge-type distributions; \cref{fig:usage_overview} reports the
citation and co-usage degree distributions; \cref{fig:temporal_split}
shows the temporal split over publication year used for
train/val/test.

\begin{table}[H]
\centering
\caption{Summary statistics of the NASA Earth Observation Knowledge
Graph (EO-KG), computed directly from the released GraphML.}
\label{tab:graph_stats}
\small
\begin{tabular}{lrr}
\toprule
\textbf{Property} & \textbf{Value} & \textbf{Note} \\
\midrule
Total nodes           & \num{150351}  & 7 distinct types \\
Total edges           & \num{436203}  & 9 relation types \\
Mean degree           & 5.80          & Across all nodes \\
Publications          & \num{138704}  & 99.5\% with title/abstract \\
Datasets              & \num{8058}    & 100\% with CMR metadata \\
Dataset temporal span & 1972--2026    & Earliest start -- latest end \\
\texttt{CITES} edges  & \num{208616}  & Citation network \\
\texttt{USES\_DATASET} edges & \num{44354} & Publication $\to$ Dataset \\
\bottomrule
\end{tabular}
\end{table}

\begin{table}[H]
\centering
\caption{Node type distribution in the NASA EO-KG with key attributes.}
\label{tab:node_types}
\small
\begin{tabular}{llrp{4.5cm}}
\toprule
\textbf{Node Type} & \textbf{Count} & \textbf{\%} & \textbf{Key Attributes} \\
\midrule
\texttt{Publication}    & \num{138704} & 92.3\% & doi, title, year, authors, abstract \\
\texttt{Dataset}        & \num{8058}   & 5.4\%  & shortName, longName, daac, cmrId, temporal extent \\
\texttt{ScienceKeyword} & \num{1609}   & 1.1\%  & name, subcategory hierarchy \\
\texttt{Instrument}     & \num{921}    & 0.6\%  & shortName, longName \\
\texttt{Platform}       & \num{455}    & 0.3\%  & shortName, longName, type \\
\texttt{Project}        & \num{415}    & 0.3\%  & shortName, longName \\
\texttt{DataCenter}     & \num{189}    & 0.1\%  & shortName, longName, url \\
\midrule
\textbf{Total}          & \num{150351} & 100\%  & \\
\bottomrule
\end{tabular}
\end{table}

\begin{table}[H]
\centering
\caption{Edge (relation) types in the NASA EO-KG with dominant source
and target node types.}
\label{tab:edge_types}
\small
\begin{tabular}{llllr}
\toprule
\textbf{Relation} & \textbf{Source} & \textbf{Target} & \textbf{Count} & \textbf{\%} \\
\midrule
\texttt{CITES}                  & \texttt{Publication}    & \texttt{Publication}    & \num{208616} & 47.8\% \\
\texttt{HAS\_APPLIEDRESEARCHAREA} & \texttt{Publication}  & \texttt{ScienceKeyword} & \num{121553} & 27.9\% \\
\texttt{USES\_DATASET}          & \texttt{Publication}    & \texttt{Dataset}        & \num{44354}  & 10.2\% \\
\texttt{HAS\_SCIENCEKEYWORD}    & \texttt{Dataset}        & \texttt{ScienceKeyword} & \num{25553}  & 5.9\% \\
\texttt{HAS\_PLATFORM}          & \texttt{Dataset}        & \texttt{Platform}       & \num{11944}  & 2.7\% \\
\texttt{HAS\_DATASET}           & \texttt{DataCenter}     & \texttt{Dataset}        & \num{11698}  & 2.7\% \\
\texttt{OF\_PROJECT}            & \texttt{Dataset}        & \texttt{Project}        & \num{8031}   & 1.8\% \\
\texttt{HAS\_INSTRUMENT}        & \texttt{Platform}       & \texttt{Instrument}     & \num{2631}   & 0.6\% \\
\texttt{HAS\_SUBCATEGORY}       & \texttt{ScienceKeyword} & \texttt{ScienceKeyword} & \num{1823}   & 0.4\% \\
\midrule
\textbf{Total}                  &                         &                         & \num{436203} & 100\% \\
\bottomrule
\end{tabular}
\end{table}

\begin{table}[H]
\centering
\caption{Distribution of NASA EO-KG datasets across Distributed Active
Archive Centers (DAACs).}
\label{tab:daac}
\small
\begin{tabular}{llrr}
\toprule
\textbf{DAAC} & \textbf{Full Name} & \textbf{Datasets} & \textbf{\%} \\
\midrule
\texttt{ASDC}     & Atmospheric Science Data Center          & \num{1601} & 19.9\% \\
\texttt{GES-DISC} & Goddard Earth Sciences DISC              & \num{1450} & 18.0\% \\
\texttt{LP DAAC}  & Land Processes DAAC                      & \num{966}  & 12.0\% \\
\texttt{NSIDC}    & Natl.\ Snow and Ice Data Center          & \num{889}  & 11.0\% \\
\texttt{PODAAC}   & Physical Oceanography DAAC               & \num{846}  & 10.5\% \\
\texttt{OB.DAAC}  & Ocean Biology DAAC                       & \num{677}  & 8.4\%  \\
\texttt{GHRC}     & Global Hydrology Resource Ctr.           & \num{639}  & 7.9\%  \\
\texttt{Other}    & Remaining DAACs                          & \num{990}  & 12.3\% \\
\midrule
\textbf{Total}    &                                          & \num{8058} & 100\%  \\
\bottomrule
\end{tabular}
\end{table}

\paragraph{Why heavy-tails matter for Stage 2.}
Panels (a)--(d) of \cref{fig:usage_overview} all exhibit heavy tails
spanning two to three orders of magnitude. This imbalance is the
practical reason we sample training negatives from
$\Pr(n) \propto \deg(n)^{0.75}$ rather than uniformly
(\cref{sec:kg-gnn}): uniform negatives would rarely include hub
datasets, yet hubs are exactly where the ranking problem is hardest
because they have many plausible partners. The edge-weight panel~(c)
additionally motivates our \emph{binary} treatment of co-usage:
weighting positives by the number of co-citing papers would upweight
already-well-known pairings at the expense of the rarer, less
well-traveled pairs that carry the most discovery value.

\paragraph{Why a temporal split.}
A random held-out split would let the model see co-usages it will
be evaluated on through common-neighbor paths (e.g., if pair
$(A, B)$ is held out but papers citing $\{A, B, C\}$ remain in
train, the training signal leaks through $C$). The temporal split
used here mitigates this leakage by partitioning pairs by the year
of each observing paper, and the strictly held-out evaluation
subset is the 3{,}944 test pairs that never appear in any
training-era paper. This mirrors the deployment scenario the
system is designed for, where the model must rank candidate
pairings that the research community has not yet explored.

\begin{table}[H]
\centering
\caption{Co-usage statistics for the NASA EO-KG. Co-usage edges are
dataset pairs cited together in at least one publication. The
train/val/test split is temporal (papers $\le 2022$ / $2023$ /
$2024$); pairs cited across multiple years can appear in more than
one split, so split sizes do not partition the unique-pair total.
\emph{Test unseen in train} is the subset of test pairs never seen
in any train-era paper.}
\label{tab:cousage_stats}
\small
\setlength{\tabcolsep}{4pt}
\begin{tabular}{lr}
\toprule
\textbf{Property} & \textbf{Value} \\
\midrule
Paper$\to$Dataset edges (total)        & \num{44354}  \\
Unique datasets cited                  & \num{2357}   \\
Datasets with $\ge 5$ papers           & \num{949}    \\
Papers citing $\ge 2$ datasets         & \num{11087}  \\
\midrule
Co-usage graph nodes                   & \num{2088}   \\
Co-usage graph edges (unique pairs)    & \num{23490}  \\
Sum of co-usage weights                & \num{64930}  \\
Max pair weight                        & \num{1051}   \\
Median / Max degree                    & 11 / \num{314} \\
\midrule
Train co-usage pairs                   & \num{13529}  \\
Val co-usage pairs                     & \num{6119}   \\
Test co-usage pairs                    & \num{6319}   \\
Test pairs unseen in train             & \num{3944}   \\
\midrule
Cold-start test pairs                  & \num{2284}   \\
Cross-DAAC test pairs                  & \num{1480}   \\
Cross-instrument test pairs            & \num{902}    \\
\bottomrule
\end{tabular}
\end{table}

\section{GNN Details, Full Baselines, and Ablation}
\label{app:gnn}

\subsection{Hyperparameters}
\label{app:gnn-hparams}
Table~\ref{tab:gnn_hparams} summarizes the full set of GNN
hyperparameters. They were not tuned per pool; the same configuration
is used for all three test pools.

\begin{table}[t]
\centering
\caption{GNN training hyperparameters for Stage~2 (shared across
GNN-Homo and GNN-Hetero).}
\label{tab:gnn_hparams}
\small
\begin{tabular}{lr}
\toprule
\textbf{Hyperparameter} & \textbf{Value} \\
\midrule
Encoder                          & 2-layer heterogeneous GraphSAGE \\
Hidden dimension                 & 128 \\
Dropout                          & 0.2 \\
Dataset node features            & SPECTER2, 768-dim (frozen) \\
Non-dataset node features        & Xavier-initialized learnable \\
Optimizer                        & Adam \\
Learning rate                    & $1 \times 10^{-3}$ \\
Max epochs                       & 300 \\
Early-stopping patience          & 30 (on val MRR) \\
Negatives per positive (train)   & 1 \\
Negative sampling distribution   & degree${}^{0.75}$ (rejection of training positives) \\
Loss                             & Binary cross-entropy with logits \\
Scorers                          & $\sigma(\mathbf{h}_i \!\cdot\! \mathbf{h}_j)$ \; or \; $\sigma(\mathrm{MLP}([\mathbf{h}_i; \mathbf{h}_j]))$ \\
Seeds                            & $\{0, 1, 2\}$ \\
\bottomrule
\end{tabular}
\end{table}

\subsection{Full baseline metrics}
\label{app:gnn-baselines-full}
\Cref{sec:kg-gnn} reports Hits@10 for compactness.
\Cref{tab:stage1_baselines} reports all five metrics
(Hits@10, Hits@50, MRR, AUC, AP) for the six Stage-1 baselines on
each of the three test pools; \cref{tab:stage2_gnn} reports the same
for GNN-Homo and GNN-Hetero with mean $\pm$ std over three seeds.

\begin{table}[t]
\centering
\caption{Stage~1 baselines on the co-usage link-prediction task (1:100 negative sampling). Higher is better for all metrics.}
\label{tab:stage1_baselines}
\small
\begin{tabular}{llrrrrr}
\toprule
\textbf{Pool} & \textbf{Baseline} & \textbf{Hits@10} & \textbf{Hits@50} & \textbf{MRR} & \textbf{AUC} & \textbf{AP} \\
\midrule
\multirow{6}{*}{\texttt{all}} 
& \texttt{Popularity}        & 0.140 & 0.562 & 0.070 & 0.542 & 0.012 \\
& \texttt{CommonNeighbors}   & 0.349 & 0.768 & 0.183 & 0.687 & 0.034 \\
& \texttt{AdamicAdar}        & 0.355 & 0.763 & 0.190 & 0.691 & 0.035 \\
& \texttt{MF-SVD}            & 0.192 & 0.433 & 0.101 & 0.491 & 0.022 \\
& \texttt{SPECTER2}          & 0.413 & 0.832 & 0.215 & 0.744 & 0.040 \\
& \texttt{BGE-base}          & 0.394 & 0.802 & 0.210 & 0.727 & 0.038 \\
\midrule
\multirow{6}{*}{\texttt{cold\_start}} 
& \texttt{Popularity}        & 0.064 & 0.297 & 0.035 & 0.371 & 0.007 \\
& \texttt{CommonNeighbors}   & 0.211 & 0.566 & 0.123 & 0.557 & 0.011 \\
& \texttt{AdamicAdar}        & 0.225 & 0.576 & 0.125 & 0.562 & 0.011 \\
& \texttt{MF-SVD}            & 0.274 & 0.503 & 0.136 & 0.538 & 0.033 \\
& \texttt{SPECTER2}          & 0.477 & 0.847 & 0.270 & 0.761 & 0.054 \\
& \texttt{BGE-base}          & 0.484 & 0.838 & 0.272 & 0.765 & 0.051 \\
\midrule
\multirow{6}{*}{\texttt{cross\_daac}} 
& \texttt{Popularity}        & 0.130 & 0.602 & 0.070 & 0.560 & 0.013 \\
& \texttt{CommonNeighbors}   & 0.228 & 0.701 & 0.122 & 0.630 & 0.018 \\
& \texttt{AdamicAdar}        & 0.235 & 0.701 & 0.128 & 0.632 & 0.019 \\
& \texttt{MF-SVD}            & 0.125 & 0.455 & 0.068 & 0.494 & 0.013 \\
& \texttt{SPECTER2}          & 0.127 & 0.683 & 0.063 & 0.605 & 0.012 \\
& \texttt{BGE-base}          & 0.103 & 0.604 & 0.053 & 0.566 & 0.011 \\
\bottomrule
\end{tabular}
\end{table}

\begin{table}[t]
\centering
\caption{Stage~2 GNN results on the co-usage link-prediction task (1:100 negative sampling, same pools as Table~\ref{tab:stage1_baselines}). Mean $\pm$ std over 3 seeds.}
\label{tab:stage2_gnn}
\small
\setlength{\tabcolsep}{3pt}
\begin{tabular}{llrrrrr}
\toprule
Variant & Pool & H@10 & H@50 & MRR & AUC & AP \\
\midrule
\texttt{GNN-Homo} & \texttt{all} & $0.448\pm0.001$ & $0.805\pm0.004$ & $0.228\pm0.005$ & $0.756\pm0.003$ & $0.042\pm0.001$ \\
\texttt{GNN-Homo} & \texttt{cold\_start} & $0.473\pm0.005$ & $0.808\pm0.006$ & $0.240\pm0.013$ & $0.758\pm0.006$ & $0.034\pm0.002$ \\
\texttt{GNN-Homo} & \texttt{cross\_daac} & $0.146\pm0.001$ & $0.594\pm0.008$ & $0.072\pm0.002$ & $0.574\pm0.004$ & $0.012\pm0.000$ \\
\midrule
\texttt{GNN-Hetero} & \texttt{all} & $0.468\pm0.003$ & $0.850\pm0.003$ & $0.225\pm0.004$ & $0.780\pm0.003$ & $0.040\pm0.001$ \\
\texttt{GNN-Hetero} & \texttt{cold\_start} & $0.519\pm0.003$ & $0.858\pm0.009$ & $0.257\pm0.004$ & $0.796\pm0.004$ & $0.040\pm0.001$ \\
\texttt{GNN-Hetero} & \texttt{cross\_daac} & $0.175\pm0.006$ & $0.669\pm0.005$ & $0.077\pm0.003$ & $0.615\pm0.004$ & $0.014\pm0.000$ \\
\bottomrule
\end{tabular}
\end{table}

\subsection{Cumulative ablation}
\label{app:gnn-ablation}
\Cref{tab:stage2p5_ablation} reports our 11-lever cumulative ablation
over both the homogeneous and heterogeneous GNN. Each row adds one
lever on top of all previously \emph{kept} levers; a lever is kept
($\checkmark$) if its addition improves mean validation MRR by more
than $10^{-3}$ over the current best, with the sole exception that the
first lever in each block (\texttt{+pubs\_uses} for hetero,
\texttt{+3\_layers} for homo) is retained as the chain's starting
point because the imported Stage-2 baseline carries no
validation-MRR record to compare against, so its checkmark does not
indicate an improvement. Three observations follow. First, the
heterogeneous baseline already dominates the homogeneous one on all
three test pools, isolating the value of type-specific message
passing. Second, after restricting publication nodes to year~$\le 2022$ to
prevent test-co-usage leakage through 2-hop message-passing paths,
adding publication-derived edges (\texttt{+pubs\_uses},
\texttt{+pubs\_area}) yields no statistically meaningful improvement
over the base heterogeneous architecture: \texttt{+pubs\_area}'s
validation-MRR delta of $+0.0009$ falls below our keep threshold
($+10^{-3}$), and Hits@10 on cold-start and cross-DAAC are slightly
\emph{lower} with publications than without. 
Third, the dot-product and MLP scorer heads form a Pareto pair on
cold-start versus cross-DAAC performance rather than a strict
ordering: dot wins cold-start (0.519) by a substantial margin, MLP
wins cross-DAAC (0.266) by an even larger one, and the two are
within $0.001$ of each other on the \texttt{all} pool. We therefore
release both embedding sets with the code so that downstream
applications can pick the head that matches their target
distribution.

\begin{table}[t]
\centering
\caption{Greedy cumulative ablation on GNN-Homo and GNN-Hetero. Starting from the Stage-2 baseline, levers are tried in the listed order; each row is the configuration after adding one lever to the previously-kept set, and a lever is marked kept (\checkmark) if it improves mean validation MRR by $>10^{-3}$ over the current best. Because the imported Stage-2 baseline carries no validation-MRR record, the first lever in each block (\texttt{+pubs\_uses} for hetero, \texttt{+3\_layers} for homo) is retained as the chain's starting point without a comparison and should not be read as an improvement---consistent with the text, publication edges do not help. The checkmarks trace this greedy search, not a single deployed stack: \texttt{+mlp\_scorer} swaps the dot head for an MLP head and \texttt{+ensemble} is a post-hoc blend, so several checked levers are mutually exclusive design choices. The retrieval feed to Agent~1 is the base heterogeneous GNN with the dot-product head and binary cross-entropy loss (\cref{sec:kg-gnn}), with the MLP head released as its Pareto complement. Hits@10, mean $\pm$ std over 3 seeds.}

\label{tab:stage2p5_ablation}
\small
\begin{tabular}{llccc}
\toprule
\textbf{Variant} & \textbf{Config} & \textbf{all} & \textbf{cold\_start} & \textbf{cross\_daac} \\
\midrule
\multirow{8}{*}{\texttt{homo}} 
& \texttt{baseline}~$\checkmark$ & 0.448{\scriptsize$\pm$0.001} & 0.473{\scriptsize$\pm$0.005} & 0.146{\scriptsize$\pm$0.001} \\
& \texttt{+3\_layers}~$\checkmark$ & 0.418{\scriptsize$\pm$0.007} & 0.418{\scriptsize$\pm$0.014} & 0.150{\scriptsize$\pm$0.019} \\
& \texttt{+residual}~$\checkmark$ & 0.450{\scriptsize$\pm$0.007} & 0.476{\scriptsize$\pm$0.009} & 0.151{\scriptsize$\pm$0.009} \\
& \texttt{+layer\_concat}~$\checkmark$ & 0.449{\scriptsize$\pm$0.007} & 0.489{\scriptsize$\pm$0.007} & 0.150{\scriptsize$\pm$0.010} \\
& \texttt{+neg\_5x}~$\checkmark$ & 0.459{\scriptsize$\pm$0.014} & 0.489{\scriptsize$\pm$0.011} & 0.149{\scriptsize$\pm$0.011} \\
& \texttt{+bpr\_loss}~ & 0.439{\scriptsize$\pm$0.004} & 0.476{\scriptsize$\pm$0.006} & 0.138{\scriptsize$\pm$0.005} \\
& \texttt{+mlp\_scorer}~ & 0.266{\scriptsize$\pm$0.001} & 0.133{\scriptsize$\pm$0.005} & 0.198{\scriptsize$\pm$0.002} \\
& \texttt{+ensemble}~$\checkmark$ & 0.466{\scriptsize$\pm$0.006} & 0.475{\scriptsize$\pm$0.025} & 0.158{\scriptsize$\pm$0.013} \\
\midrule
\multirow{12}{*}{\texttt{hetero}} 
& \texttt{baseline}~$\checkmark$ & 0.468{\scriptsize$\pm$0.003} & 0.519{\scriptsize$\pm$0.003} & 0.175{\scriptsize$\pm$0.006} \\
& \texttt{+pubs\_uses}~$\checkmark$ & 0.467{\scriptsize$\pm$0.003} & 0.513{\scriptsize$\pm$0.011} & 0.171{\scriptsize$\pm$0.014} \\
& \texttt{+pubs\_cites}~ & 0.458{\scriptsize$\pm$0.005} & 0.509{\scriptsize$\pm$0.017} & 0.163{\scriptsize$\pm$0.009} \\
& \texttt{+pubs\_area}~ & 0.473{\scriptsize$\pm$0.008} & 0.516{\scriptsize$\pm$0.009} & 0.170{\scriptsize$\pm$0.011} \\
& \texttt{+edge\_weights}~ & 0.467{\scriptsize$\pm$0.003} & 0.513{\scriptsize$\pm$0.011} & 0.171{\scriptsize$\pm$0.014} \\
& \texttt{+3\_layers}~ & 0.452{\scriptsize$\pm$0.007} & 0.511{\scriptsize$\pm$0.009} & 0.155{\scriptsize$\pm$0.016} \\
& \texttt{+residual}~ & 0.462{\scriptsize$\pm$0.003} & 0.500{\scriptsize$\pm$0.018} & 0.173{\scriptsize$\pm$0.005} \\
& \texttt{+layer\_concat}~ & 0.464{\scriptsize$\pm$0.002} & 0.515{\scriptsize$\pm$0.007} & 0.165{\scriptsize$\pm$0.004} \\
& \texttt{+neg\_5x}~ & 0.469{\scriptsize$\pm$0.005} & 0.497{\scriptsize$\pm$0.003} & 0.192{\scriptsize$\pm$0.012} \\
& \texttt{+bpr\_loss}~$\checkmark$ & 0.465{\scriptsize$\pm$0.002} & 0.508{\scriptsize$\pm$0.007} & 0.170{\scriptsize$\pm$0.013} \\
& \texttt{+mlp\_scorer}~$\checkmark$ & 0.467{\scriptsize$\pm$0.005} & 0.389{\scriptsize$\pm$0.006} & 0.266{\scriptsize$\pm$0.011} \\
& \texttt{+ensemble}~$\checkmark$ & 0.470{\scriptsize$\pm$0.008} & 0.377{\scriptsize$\pm$0.014} & 0.271{\scriptsize$\pm$0.017} \\
\bottomrule
\end{tabular}
\end{table}

\section{Three-Agent Prompts and JSON Schemas}
\label{app:prompts}

The four colored boxes below reproduce verbatim the system prompts,
user-message templates, and JSON output schemas used in the
three-agent pipeline of~\cref{sec:pipeline}. The same prompts are
used regardless of which backbone (GPT-5.2 or Claude Sonnet~4.6) is
playing the role, so that any difference in the resulting scores
reflects the backbone, not the prompt. All four call sites use
temperature $0$ and the API's strict-JSON response format
(GPT: \texttt{response\_format=json\_object}; Claude: defensive
post-hoc fence stripping).

\begin{tcolorbox}[
  enhanced, breakable,
  colback=blue!4, colframe=blue!55!black, colbacktitle=blue!55!black,
  coltitle=white, fonttitle=\bfseries,
  title={Agent 1 — Pair-Level Filter},
  arc=2pt, boxrule=0.6pt, left=4pt, right=4pt, top=3pt, bottom=3pt,
]
\textbf{Backbone:} GPT-5.2 \emph{or} Claude Sonnet 4.6. \quad
\textbf{Temperature:} $0$. \quad
\textbf{Max output tokens:} 400.

\medskip
\textbf{System prompt}\par\smallskip
\small\itshape
You are an expert Earth-science research scientist. Your job is to
evaluate whether two Earth-observation datasets could plausibly be
combined in a single scientific study. Base your judgment on the
datasets' scientific content as described in their abstracts, not on
whether you have seen them combined before in published work.
Respond with JSON only, in the exact schema requested.
\normalfont\upshape

\medskip
\textbf{User-message template}\par\smallskip
\small\ttfamily
Please evaluate the following pair of Earth-observation datasets.\\[2pt]
Dataset A:\\
\hspace*{1em}Short name: \{sn\_a\}\\
\hspace*{1em}Long name:\ \{ln\_a\}\\
\hspace*{1em}Abstract:\ \ \{abs\_a\}\\[2pt]
Dataset B:\\
\hspace*{1em}Short name: \{sn\_b\}\\
\hspace*{1em}Long name:\ \{ln\_b\}\\
\hspace*{1em}Abstract:\ \ \{abs\_b\}\\[2pt]
Rate this pairing on two axes:\\
1. SCIENTIFIC PLAUSIBILITY (1--5): Could a research team reasonably\\
\hspace*{1em}combine these datasets in a single scientific study?\\
2. NOVELTY (1--5): How novel or non-obvious is this combination?
\normalfont

\medskip
\textbf{Output JSON schema}\par\smallskip
\small\ttfamily
\{ "plausibility": <1--5 int>,\\
\hspace*{1em}"novelty":\ \ \ \ \ \ <1--5 int>,\\
\hspace*{1em}"rationale":\ \ \ \ "<2--3 sentence explanation>" \}
\normalfont

\medskip
\textbf{Inputs:} \texttt{shortName}, \texttt{longName}, and abstract
(truncated to 1{,}200 characters) for each dataset. \quad
\textbf{Output:} two integer scores plus a free-text rationale.
\end{tcolorbox}

\begin{tcolorbox}[
  enhanced, breakable,
  colback=green!4, colframe=green!50!black, colbacktitle=green!50!black,
  coltitle=white, fonttitle=\bfseries,
  title={Agent 2 — Hypothesis Generator},
  arc=2pt, boxrule=0.6pt, left=4pt, right=4pt, top=3pt, bottom=3pt,
]
\textbf{Backbone:} GPT-5.2 \emph{or} Claude Sonnet 4.6. \quad
\textbf{Temperature:} $0$. \quad
\textbf{Max output tokens:} 900.

\medskip
\textbf{System prompt}\par\smallskip
\small\itshape
You are an expert Earth-observation research scientist. You will be
given two NASA datasets. Generate ONE concrete, publishable research
hypothesis that could be tested by combining them. Focus on
scientific content, not data-availability caveats. Respond with
JSON only.
\normalfont\upshape

\medskip
\textbf{User-message template}\par\smallskip
\small\ttfamily
Dataset A:\\
\hspace*{1em}Short name: \{sn\_a\}\\
\hspace*{1em}Long name:\ \{ln\_a\}\\
\hspace*{1em}Abstract:\ \ \{abs\_a\}\\[2pt]
Dataset B:\\
\hspace*{1em}Short name: \{sn\_b\}\\
\hspace*{1em}Long name:\ \{ln\_b\}\\
\hspace*{1em}Abstract:\ \ \{abs\_b\}\\[2pt]
Generate one specific, testable research hypothesis that combines\\
these two datasets. The hypothesis must be concrete enough that a\\
team could plan an actual study, not a vague ``one could investigate\\
X'' statement.
\normalfont

\medskip
\textbf{Output JSON schema}\par\smallskip
\small\ttfamily
\{ "research\_question":\ \ \ \ \ "<one-sentence question>",\\
\hspace*{1em}"hypothesis":\ \ \ \ \ \ \ \ \ \ \ \ "<testable hypothesis, 1--2 sent.>",\\
\hspace*{1em}"analysis\_method":\ \ \ \ \ \ "<how datasets test it, 1--2 sent.>",\\
\hspace*{1em}"expected\_finding":\ \ \ \ \ "<what would support, 1 sent.>",\\
\hspace*{1em}"scientific\_importance": "<why it matters, 1--2 sent.>",\\
\hspace*{1em}"domain":\ \ \ \ \ \ \ \ \ \ \ \ \ \ \ \ "<primary domain, 1--3 words>" \}
\normalfont

\medskip
\textbf{Inputs:} the surviving pair $(d_i, d_j)$ from the tiered
top-40 selection, with the same dataset metadata Agent~1 received.
Crucially, Agent~2 does \emph{not} see Agent~1's scores.\quad
\textbf{Output:} a six-field structured hypothesis.
\end{tcolorbox}

\begin{tcolorbox}[
  enhanced, breakable,
  colback=orange!5, colframe=orange!70!black, colbacktitle=orange!70!black,
  coltitle=white, fonttitle=\bfseries,
  title={Agent 3 — Judge (Blind condition)},
  arc=2pt, boxrule=0.6pt, left=4pt, right=4pt, top=3pt, bottom=3pt,
]
\textbf{Backbone:} GPT-5.2 \emph{or} Claude Sonnet 4.6. \quad
\textbf{Temperature:} $0$. \quad
\textbf{Max output tokens:} 500.

\medskip
\textbf{System prompt}\par\smallskip
\small\itshape
You are a senior Earth-observation research scientist reviewing a
proposed research hypothesis. Judge the hypothesis on its scientific
merit alone; you do not need to verify data availability. Respond
with JSON only.
\normalfont\upshape

\medskip
\textbf{User-message template}\par\smallskip
\small\ttfamily
Here is a proposed Earth-science research hypothesis:\\[2pt]
Research question:\ \ \ \ \{research\_question\}\\
Hypothesis:\ \ \ \ \ \ \ \ \ \ \ \{hypothesis\}\\
Proposed method:\ \ \ \ \ \ \{analysis\_method\}\\
Expected finding:\ \ \ \ \{expected\_finding\}\\
Scientific importance: \{scientific\_importance\}\\
Domain:\ \ \ \ \ \ \ \ \ \ \ \ \ \ \ \{domain\}\\[2pt]
Rate 1--5:\\
\hspace*{1em}IMPORTANCE:\ \ \ 1 = trivial,\ 5 = meaningfully advances field\\
\hspace*{1em}TRACTABILITY: 1 = needs breakthroughs,\ 5 = ready to execute\\
\hspace*{1em}NOVELTY:\ \ \ \ \ \ 1 = already well-studied,\ 5 = opens new line
\normalfont

\medskip
\textbf{Output JSON schema}\par\smallskip
\small\ttfamily
\{ "importance":\ \ <1--5>,\ "tractability": <1--5>,\\
\hspace*{1em}"novelty":\ \ \ \ \ <1--5>,\ "rationale":\ \ \ \ "<2--3 sentences>" \}
\normalfont

\medskip
\textbf{Inputs:} only the six hypothesis fields produced by Agent~2;
no dataset metadata. \quad
\textbf{Output:} three integer scores plus rationale.
\end{tcolorbox}

\begin{tcolorbox}[
  enhanced, breakable,
  colback=red!4, colframe=red!55!black, colbacktitle=red!55!black,
  coltitle=white, fonttitle=\bfseries,
  title={Agent 3 — Judge (Contextual condition)},
  arc=2pt, boxrule=0.6pt, left=4pt, right=4pt, top=3pt, bottom=3pt,
]
\textbf{Backbone:} GPT-5.2 \emph{or} Claude Sonnet 4.6. \quad
\textbf{Temperature:} $0$. \quad
\textbf{Max output tokens:} 500.

\medskip
\textbf{System prompt}\par\smallskip
\small\itshape
You are a senior Earth-observation research scientist reviewing a
proposed research hypothesis along with the datasets it uses. Judge
the hypothesis on scientific merit and the appropriateness of the
dataset combination. Respond with JSON only.
\normalfont\upshape

\medskip
\textbf{User-message template}\par\smallskip
\small\ttfamily
Dataset A:\\
\hspace*{1em}Short name: \{sn\_a\}\\
\hspace*{1em}Long name:\ \{ln\_a\}\\
\hspace*{1em}Abstract:\ \ \{abs\_a\}\\[2pt]
Dataset B:\\
\hspace*{1em}Short name: \{sn\_b\}\\
\hspace*{1em}Long name:\ \{ln\_b\}\\
\hspace*{1em}Abstract:\ \ \{abs\_b\}\\[2pt]
Proposed research hypothesis:\\[2pt]
Research question:\ \ \ \ \{research\_question\}\\
Hypothesis:\ \ \ \ \ \ \ \ \ \ \ \{hypothesis\}\\
Proposed method:\ \ \ \ \ \ \{analysis\_method\}\\
Expected finding:\ \ \ \ \{expected\_finding\}\\
Scientific importance: \{scientific\_importance\}\\
Domain:\ \ \ \ \ \ \ \ \ \ \ \ \ \ \ \{domain\}\\[2pt]
Rate 1--5 (same scales as the blind condition).
\normalfont

\medskip
\textbf{Output JSON schema}\par\smallskip
\small\ttfamily
\{ "importance":\ \ <1--5>,\ "tractability": <1--5>,\\
\hspace*{1em}"novelty":\ \ \ \ \ <1--5>,\ "rationale":\ \ \ \ "<2--3 sentences>" \}
\normalfont

\medskip
\textbf{Inputs:} the same six hypothesis fields as the blind judge,
\emph{plus} the underlying pair's \texttt{shortName},
\texttt{longName}, and abstracts (truncated to 1{,}200 characters).
\quad
\textbf{Output:} three integer scores plus rationale.
\end{tcolorbox}

\section{Factorial Design: Extended Analyses}
\label{app:factorial}

This appendix presents three extended views of the Stage-5 factorial
results summarized in \cref{sec:results}. 

\paragraph{Inter-rater agreement (Table~\ref{tab:inter_rater}).}
\Cref{tab:inter_rater} reports the full inter-rater statistics
referenced in \cref{sec:results}: Pearson $r$, Spearman $\rho$,
Cohen's quadratic-weighted $\kappa_q$, and the exact-match
percentage between GPT-5.2 and Claude Sonnet~4.6 on identical
hypotheses, separated by judging condition. The headline ordering
holds for every metric (tractability $>$ novelty $>$ importance
under both conditions), and the contextual-condition gain on
importance ($\kappa_q$: $0.14 \to 0.24$, exact match: $37\% \to
63\%$) is the largest single improvement in the table.

\begin{table}[H]
\centering
\caption{Inter-rater agreement between GPT-5.2 and Claude Sonnet 4.6
on identical hypotheses. Judges agree best on tractability, less on
novelty, and only weakly on importance. Showing dataset context
improves agreement on all three axes; the largest gain is on
importance ($\kappa_q$: $0.14 \to 0.24$, exact match: $37\% \to
63\%$). $\kappa_q$ is Cohen's quadratic-weighted kappa.}
\label{tab:inter_rater}
\small
\setlength{\tabcolsep}{4pt}
\begin{tabular}{llccccc}
\toprule
Axis & Cond.\ & $n$ & Pearson $r$ & Spearman $\rho$ & $\kappa_q$ & Exact \% \\
\midrule
Importance   & blind & 160 & +0.351 & +0.337 & +0.137 & 37 \\
Importance   & ctx   & 160 & +0.397 & +0.366 & +0.241 & 63 \\
Tractability & blind & 160 & +0.627 & +0.578 & +0.472 & 54 \\
Tractability & ctx   & 160 & +0.601 & +0.585 & +0.572 & 58 \\
Novelty      & blind & 160 & +0.487 & +0.456 & +0.356 & 57 \\
Novelty      & ctx   & 160 & +0.527 & +0.541 & +0.512 & 71 \\
\bottomrule
\end{tabular}
\end{table}

\begin{figure}[t]
\centering
\includegraphics[width=\linewidth]{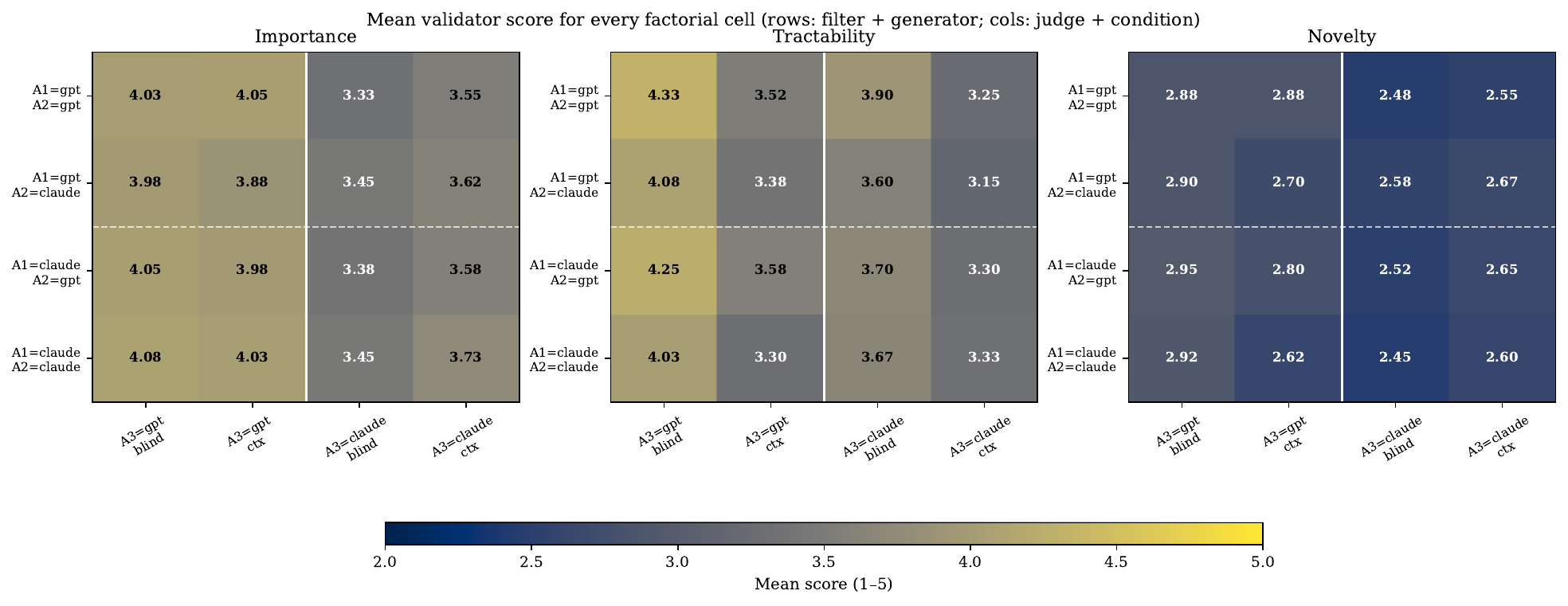}
\caption{Per-cell mean validator score for each axis (importance,
tractability, novelty). Rows: $(a_1, a_2)$ filter--generator
combinations. Columns: $(a_3, \text{condition})$ judge--condition
combinations. Color scale uses the cividis colormap on the score
range $[2.0, 5.0]$. The vertical white divider separates the GPT
and Claude judges; the horizontal dashed divider separates GPT and
Claude filters.}
\label{fig:factorial_heatmap}
\end{figure}

\paragraph{Per-cell mean scores (Figure~\ref{fig:factorial_heatmap}).}
Three heatmaps report the mean validator score for every cell of
the $2{\times}2{\times}2$ factorial, with the blind/contextual
condition appearing as adjacent column-pairs. Rows are
$(a_1, a_2)$ filter--generator combinations; columns are
$(a_3, \text{condition})$. The visual signature of the variance
decomposition (\cref{fig:factorial_variance}) is immediately
visible: \textbf{importance} (left panel) splits cleanly by judge---
the two GPT-judge columns are uniformly darker than the two Claude-
judge columns regardless of $(a_1, a_2)$, with mean shifts of
$\sim 0.5$--$0.7$ points; \textbf{tractability} (middle panel)
splits by condition---blind columns are uniformly darker than ctx
columns within each judge; \textbf{novelty} (right panel) is mostly
flat, consistent with its low total variance. The white vertical
divider separates GPT-judge from Claude-judge columns; the dashed
horizontal divider separates GPT-filter from Claude-filter rows.

\begin{figure}[t]
\centering
\includegraphics[width=\linewidth]{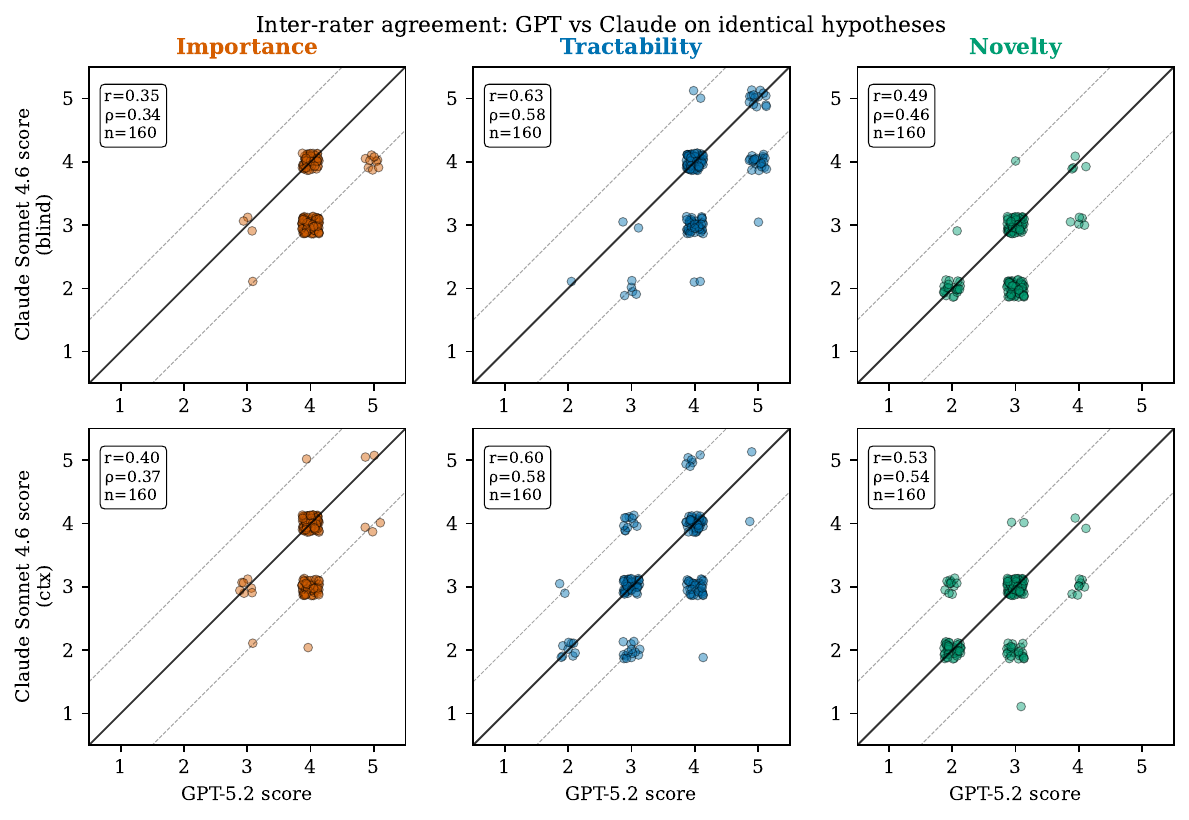}
\caption{Inter-rater scatter: GPT-5.2 score (x) vs.\ Claude Sonnet
4.6 score (y) for each hypothesis under each judging condition.
Top row: blind condition. Bottom row: contextual condition.
Columns: importance, tractability, novelty. Solid line: equality;
dashed lines: $\pm 1$ tolerance. Pearson $r$, Spearman $\rho$, and
sample size annotated per panel. Tractability shows the highest
agreement; importance shows the lowest, with a visible upward bias
for GPT that is partially attenuated by adding dataset context.}
\label{fig:factorial_agreement}
\end{figure}

\paragraph{Inter-rater scatter (Figure~\ref{fig:factorial_agreement}).}
For every hypothesis judged under both backbones in the same
condition, the scatter plots GPT's score (x-axis) against Claude's
score (y-axis), with jitter added to separate ties. The diagonal is
perfect agreement; the dashed diagonals mark $\pm 1$ tolerance.
Each panel reports Pearson $r$ and Spearman $\rho$ over its
160-point sample. The visualization makes the pattern from
\cref{tab:inter_rater} concrete: tractability scatter clusters
tightly around the diagonal under both conditions; novelty scatter
is moderately tight; importance scatter exhibits a visible vertical
bias (GPT systematically high) that narrows but does not vanish in
the contextual condition.

\section{Flagship Hypotheses}
\label{app:flagship}

This appendix lists the five flagship hypotheses referenced in
\cref{sec:results}. Selection criterion: the highest combined
importance + tractability scores with a maximum cross-judge
disagreement of one on every axis, i.e., consensus high quality
across all four judge conditions
($a_3 \!\in\! \{\text{GPT}, \text{Claude}\} \!\times\!
\{\text{blind}, \text{contextual}\}$). The score columns
\emph{Imp}, \emph{Tract}, and \emph{Novel} report the mean across
those four judgments; \emph{Max~$\Delta$} is the largest
disagreement on any axis.

\begin{table*}[h]
\centering
\small
\caption{Five flagship hypotheses from the 2$\times$2$\times$2 factorial. Each was generated by the (Agent 1, Agent 2) combination listed, then rated by both judges under both conditions (blind and contextual); the \textit{Imp}, \textit{Tract}, and \textit{Novel} columns report the mean across these 4 validator judgments. \textit{Max $\Delta$} is the largest disagreement on any axis among the 4 judgments; a low value indicates the hypothesis is rated consistently across agents and conditions.}
\label{tab:flagship}
\begin{tabular}{p{2.3cm}p{2.1cm}cccccp{4.2cm}}
\toprule
Dataset pair & Domain & Imp & Tract & Novel & Max $\Delta$ & Source (A1→A2) & Research question (abbreviated) \\
\midrule
SPL4SMGP + MYD13Q1 & Ecohydrology & 4.00 & 4.50 & 3.00 & 1 & Claude$\to$GPT & Does sub-seasonal root-zone soil moisture variability control the timing and magnitude of vegetation greeni… \\
IRKUB1B + IRMCR1B & glaciology & 4.50 & 3.75 & 3.25 & 1 & GPT$\to$GPT & Do along-track changes in Ku-band surface echo strength predict the presence and thickness of near-surface… \\
MOD08\_M3 + CAL\_LID\_L3\_Tropospheric\_APro\_CloudFree-Standard-V4-20 & Aerosol-cloud & 4.50 & 3.50 & 3.25 & 1 & Claude$\to$GPT & Does the vertical placement of tropospheric aerosol (near-surface vs elevated layers) modulate the monthly… \\
MYD13A2 + MYD15A2H & vegetation phenology & 4.00 & 4.25 & 3.00 & 1 & GPT$\to$GPT & Does the seasonal relationship between canopy greenness (EVI/NDVI) and canopy structure (LAI/FPAR) exhibit… \\
MOD06\_L2 + CAL\_LID\_L2\_VFM-Standard-V4-20 & Cloud remote sensing & 4.00 & 4.75 & 2.50 & 1 & GPT$\to$GPT & Do MODIS/Terra MOD06\_L2 cloud optical thickness and effective radius retrievals exhibit systematic biases w… \\
\bottomrule
\end{tabular}
\end{table*}

\paragraph{Qualitative reading.}
\Cref{tab:flagship} surfaces three properties worth noting. First,
the flagship pairs mix same-mission, multi-sensor or multi-level
combinations (e.g., IceBridge Ku-band radar with MCoRDS, or
Aqua-MODIS EVI with Aqua-MODIS LAI) with cross-mission,
cross-archive combinations whose measurements are nonetheless
physically co-located (e.g., Terra-MODIS aerosol with CALIPSO lidar,
or SMAP soil moisture with Aqua-MODIS vegetation indices),
suggesting that the ranker prefers pairings that are jointly
co-registerable and physically complementary rather than merely
semantically similar. Second, the flagship set spans five disjoint scientific domains
(ecohydrology, glaciology, aerosol--cloud interactions,
vegetation phenology, and methodological bias correction in cloud
retrieval), indicating that the pipeline does not collapse onto a
single theme even when optimizing only for importance and
tractability. Third, the \emph{novelty} column for every flagship
hypothesis sits in the $2.5$--$3.25$ range, the boundary between
``moderately novel'' and ``opens a new line of inquiry''---the
rating-surface region where the tiered top-40 selection of
\cref{sec:pipeline} concentrates effort. The flagship set is
therefore a calibrated sample of what the pipeline considers
ambitious-but-plausible, not a curated highlight reel.

\paragraph{Full reproductions.}
\Cref{app:qualitative} reproduces three hypotheses in their
entirety as gray chatboxes spanning the score spectrum, including
the flagship glaciology entry (\textit{IRKUB1B $\times$ IRMCR1B}) in
its complete six-field form together with all four Agent~3
verdicts. The released hypothesis-corpus TSV files
(\texttt{hypotheses\_\{a1\}\_\{a2\}.tsv}) contain the unabridged
JSON outputs for all 160 hypotheses in the factorial.

\section{Qualitative Hypothesis Examples}
\label{app:qualitative}

To complement the aggregate statistics of \cref{sec:results} and the
flagship table of \cref{app:flagship}, we reproduce three full
hypotheses spanning the score spectrum: a top-quartile case
(\textsc{High}, mean combined score $11.50$), a near-median case
(\textsc{Medium}, $10.25$), and a bottom-decile case
(\textsc{Low}, $7.25$). Each box reports the dataset pair as fed to
Agent~2, the structured six-field hypothesis Agent~2 returned, and
the four Agent~3 verdicts (judge $\in \{$GPT, Claude$\}$ $\times$
condition $\in \{$blind, contextual$\}$). Selected judge rationales
are quoted to make the cross-judge calibration shift visible at the
example level.

\begin{tcolorbox}[
  enhanced, breakable,
  colback=gray!8, colframe=gray!55!black, colbacktitle=gray!55!black,
  coltitle=white, fonttitle=\bfseries,
  title={Example 1 (\textsc{High}) --- IRKUB1B $\times$ IRMCR1B \,(glaciology)},
  arc=2pt, boxrule=0.6pt, left=4pt, right=4pt, top=3pt, bottom=3pt,
]
\textbf{Filter $\to$ Generator:} GPT $\to$ GPT \quad
\textbf{Source stratum:} A (predicted novel).

\medskip
\textbf{Dataset pair}\par\smallskip
\small
\textbf{IRKUB1B} --- IceBridge Ku-Band Radar L1B Geolocated Radar
Echo Strength Profiles V002 (NSIDC).\\
\textbf{IRMCR1B} --- IceBridge MCoRDS L1B Geolocated Radar Echo
Strength Profiles V002 (NSIDC).
\normalsize

\medskip
\textbf{Agent 2 — generated hypothesis}\par\smallskip
\small
\textbf{Research question.} Do along-track changes in Ku-band
surface echo strength predict the presence and thickness of
near-surface firn aquifers and/or ice lenses detected by MCoRDS in
Greenland's percolation zone?\par\smallskip
\textbf{Hypothesis.} Where MCoRDS echograms show a laterally
continuous, high-reflectivity near-surface horizon (ice lens) or a
strong diffuse-scattering zone consistent with a firn aquifer, the
Ku-band radar altimeter exhibits a systematic increase in surface
echo strength and a coincident shift in leading-edge/peak-power
range (apparent elevation bias) relative to adjacent dry-firn
segments; the magnitude of the Ku-band anomaly scales with the
MCoRDS-inferred thickness of the wet/ice-rich layer.\par\smallskip
\textbf{Analysis method.} Co-register IRKUB1B and IRMCR1B profiles
along overlapping flight lines; classify segments into dry-firn,
ice-lens, or firn-aquifer/wet-firn using MCoRDS reflectors;
estimate layer depth from MCoRDS two-way travel time; test
class-conditional differences in Ku-band echo strength and
waveform-derived range, controlling for surface slope and
roughness proxies.\par\smallskip
\textbf{Expected finding.} Repeatable Ku-band echo-strength
increases and range/leading-edge shifts over MCoRDS-identified
wet/ice-rich layers, with anomaly magnitude positively correlated
with MCoRDS-inferred layer thickness.\par\smallskip
\textbf{Scientific importance.} Linking Ku-band surface-return
behavior to subsurface hydrologic state enables bias correction of
radar-altimetry-derived elevation change in melt-affected regions
and provides an altimetry-only route to map firn aquifers and ice
lenses, sharpening Greenland mass-balance estimates.
\normalsize

\medskip
\textbf{Agent 3 verdicts (Importance / Tractability / Novelty, 1--5)}
\par\smallskip
\small
\setlength{\tabcolsep}{6pt}
\begin{tabular}{lccc}
\toprule
Judge \,(condition) & Imp & Tract & Nov \\
\midrule
GPT-5.2 \,(blind)         & 5 & 4 & 3 \\
GPT-5.2 \,(contextual)    & 5 & 4 & 4 \\
Claude 4.6 \,(blind)      & 4 & 4 & 3 \\
Claude 4.6 \,(contextual) & 4 & 3 & 3 \\
\midrule
\textbf{Mean across judges} & \textbf{4.50} & \textbf{3.75} & \textbf{3.25} \\
\bottomrule
\end{tabular}
\normalsize

\medskip
\textbf{Selected rationales}\par\smallskip
\small\itshape
\textbf{GPT (blind):} ``Establishing a quantitative link between
Ku-band waveform/echo-strength anomalies and near-surface wet/ice-
rich firn structures would directly improve elevation-change
interpretation and bias correction.''\par\smallskip
\textbf{Claude (contextual):} ``Quantifying how near-surface
wet/ice-rich layers bias Ku-band altimetry returns is directly
relevant to correcting CryoSat-2 and Sentinel-6 elevation-change
records in Greenland's percolation zone.''
\normalfont\upshape
\end{tcolorbox}

\begin{tcolorbox}[
  enhanced, breakable,
  colback=gray!8, colframe=gray!55!black, colbacktitle=gray!55!black,
  coltitle=white, fonttitle=\bfseries,
  title={Example 2 (\textsc{Medium}) --- OMGLER $\times$ OMUVBG \,(UV radiation)},
  arc=2pt, boxrule=0.6pt, left=4pt, right=4pt, top=3pt, bottom=3pt,
]
\textbf{Filter $\to$ Generator:} Claude $\to$ GPT \quad
\textbf{Source stratum:} A (predicted novel).

\medskip
\textbf{Dataset pair}\par\smallskip
\small
\textbf{OMGLER} --- OMI/Aura Global Geometry-Dependent Surface LER
1-Orbit L2 Swath 13$\times$24~km V3 (GES-DISC).\\
\textbf{OMUVBG} --- OMI/Aura Surface UVB Irradiance and Erythemal
Dose Daily L2 Global Gridded $0.25^\circ$ V3 (GES-DISC).
\normalsize

\medskip
\textbf{Agent 2 — generated hypothesis}\par\smallskip
\small
\textbf{Research question.} How much of the day-to-day variability
in OMI-derived surface UVB irradiance and erythemal dose can be
explained by geometry-dependent surface reflectance (GLER)
variations observed by OMI?\par\smallskip
\textbf{Hypothesis.} After controlling for solar zenith angle and
cloud screening, higher OMGLER within a $0.25^\circ$ grid cell is
associated with systematically higher OMUVBG surface UVB
irradiance and erythemal dose on the same day, with strongest
sensitivity over snow/ice and arid regions; the GLER--UVB
relationship is nonlinear, saturating at very high GLER due to
multiple-scattering effects.\par\smallskip
\textbf{Analysis method.} Collocate OMGLER swath pixels to OMUVBG
$0.25^\circ$ daily cells; compute per-cell daily statistics of
GLER alongside OMUVBG using high-quality, low-cloud candidates;
fit a mixed-effects or generalized additive model predicting
OMUVBG from GLER while adjusting for solar zenith angle, viewing
geometry, season, and latitude; test interaction terms for surface
type proxies.\par\smallskip
\textbf{Expected finding.} A statistically significant positive
GLER coefficient under comparable illumination and cloud
conditions, with the largest effect sizes in high-GLER environments.
\par\smallskip
\textbf{Scientific importance.} Quantifying the sensitivity of
surface UV exposure estimates to geometry-dependent reflectance
improves bias correction of satellite UV products over bright
surfaces, with direct relevance to UV climatology and public-
health exposure assessment.
\normalsize

\medskip
\textbf{Agent 3 verdicts (Importance / Tractability / Novelty, 1--5)}
\par\smallskip
\small
\setlength{\tabcolsep}{6pt}
\begin{tabular}{lccc}
\toprule
Judge \,(condition) & Imp & Tract & Nov \\
\midrule
GPT-5.2 \,(blind)         & 4 & 5 & 3 \\
GPT-5.2 \,(contextual)    & 4 & 4 & 3 \\
Claude 4.6 \,(blind)      & 3 & 4 & 2 \\
Claude 4.6 \,(contextual) & 3 & 4 & 2 \\
\midrule
\textbf{Mean across judges} & \textbf{3.50} & \textbf{4.25} & \textbf{2.50} \\
\bottomrule
\end{tabular}
\normalsize

\medskip
\textbf{Selected rationales}\par\smallskip
\small\itshape
\textbf{GPT (blind):} ``The hypothesis targets a known but
consequential error source in satellite UV products---surface
reflectance/BRDF effects---so quantifying day-to-day GLER-driven
variability can materially improve interpretation.''\par\smallskip
\textbf{Claude (blind):} ``The influence of surface reflectance on
UV irradiance is physically well-established and has been studied
extensively in the context of OMI and other satellite UV
retrievals, including snow/ice albedo \dots [novelty=2].''
\normalfont\upshape

\medskip
\small
\emph{Cross-judge note.} This example illustrates the calibration
shift quantified in \cref{sec:results}: GPT and Claude both rate
tractability highly and agree on the analysis design, but GPT
returns importance$=4$ and novelty$=3$ where Claude returns
importance$=3$ and novelty$=2$. The hypothesis is sound; only its
absolute level depends on which backbone is asked.
\normalsize
\end{tcolorbox}

\begin{tcolorbox}[
  enhanced, breakable,
  colback=gray!8, colframe=gray!55!black, colbacktitle=gray!55!black,
  coltitle=white, fonttitle=\bfseries,
  title={Example 3 (\textsc{Low}) --- OMOCLO $\times$ ML2CH3CL \,(stratospheric chemistry)},
  arc=2pt, boxrule=0.6pt, left=4pt, right=4pt, top=3pt, bottom=3pt,
]
\textbf{Filter $\to$ Generator:} GPT $\to$ GPT \quad
\textbf{Source stratum:} A (predicted novel).

\medskip
\textbf{Dataset pair}\par\smallskip
\small
\textbf{OMOCLO} --- OMI/Aura Chlorine Dioxide (OClO) Total Column
1-orbit L2 Swath 13$\times$24~km V003 (GES-DISC).\\
\textbf{ML2CH3CL} --- MLS/Aura L2 Methyl Chloride (CH$_3$Cl) Mixing
Ratio V005 (GES-DISC).
\normalsize

\medskip
\textbf{Agent 2 — generated hypothesis}\par\smallskip
\small
\textbf{Research question.} Do enhancements in lower-stratospheric
methyl chloride (CH$_3$Cl) precede and predict increases in polar
chlorine dioxide (OClO) total columns during the onset of Antarctic
springtime chlorine activation?\par\smallskip
\textbf{Hypothesis.} During late winter to early spring (Aug--Sep)
in the Antarctic vortex (60--80$^\circ$S), positive anomalies in
MLS CH$_3$Cl mixing ratios at 100--30~hPa are followed within
0--14 days by statistically significant increases in OMI OClO
total columns under comparable solar-zenith-angle conditions,
with the strongest effect in colder-vortex years.\par\smallskip
\textbf{Analysis method.} Collocate daily OMOCLO OClO columns with
MLS CH$_3$Cl profiles averaged over the same vortex-defined air
mass (equivalent latitude/PV filtering); compute CH$_3$Cl
anomalies at 100--30~hPa; apply lagged regression and event-based
composites to test whether CH$_3$Cl anomalies lead OClO increases,
controlling for SZA and stratifying by vortex temperature.
\par\smallskip
\textbf{Expected finding.} A robust positive lagged association in
which elevated CH$_3$Cl precedes higher OClO columns by up to
$\sim$2 weeks, larger in colder-vortex years.\par\smallskip
\textbf{Scientific importance.} Linking an organic-chlorine tracer
to observed chlorine activation would clarify how variability in
chlorine source-gas transport modulates polar halogen chemistry
and ozone-loss potential.
\normalsize

\medskip
\textbf{Agent 3 verdicts (Importance / Tractability / Novelty, 1--5)}
\par\smallskip
\small
\setlength{\tabcolsep}{6pt}
\begin{tabular}{lccc}
\toprule
Judge \,(condition) & Imp & Tract & Nov \\
\midrule
GPT-5.2 \,(blind)         & 3 & 4 & 3 \\
GPT-5.2 \,(contextual)    & 3 & 2 & 3 \\
Claude 4.6 \,(blind)      & 2 & 2 & 2 \\
Claude 4.6 \,(contextual) & 2 & 2 & 1 \\
\midrule
\textbf{Mean across judges} & \textbf{2.50} & \textbf{2.50} & \textbf{2.25} \\
\bottomrule
\end{tabular}
\normalsize

\medskip
\textbf{Selected rationales}\par\smallskip
\small\itshape
\textbf{Claude (blind):} ``The fundamental mechanisms linking
chlorine source gases to polar chlorine activation and OClO
formation are well-established; attributing interannual OClO
variability to CH$_3$Cl transport anomalies specifically is
unlikely to yield a strong signal.''\par\smallskip
\textbf{GPT (contextual):} ``CH$_3$Cl is a weak and indirect proxy
for inorganic chlorine available for activation; the proposed
lagged relationship is plausible in principle but execution is
difficult given measurement uncertainty.''
\normalfont\upshape

\medskip
\small
\emph{Failure mode.} The two judges concur that the underlying
chemistry is well-studied and that CH$_3$Cl is a weak proxy for
the actual activation pathway; the GPT--Claude tractability gap
under the blind condition (4 vs.\ 2) reflects different
priors about what counts as an executable observational study
when the predicted effect size is small.
\normalsize
\end{tcolorbox}


\end{document}